\documentclass{article} %
\usepackage{iclr2027_conference,times}

\usepackage{hyperref}
\usepackage{url}

\usepackage{graphicx}
\usepackage{multirow}
\usepackage{hyperref}       %
\usepackage{url}            %
\usepackage{booktabs}       %
\usepackage{amsmath}
\usepackage{amssymb}
\usepackage{bm}
\usepackage{amsthm}
\usepackage{upgreek}
\usepackage{xr}

\makeatletter
\newcommand*{\addFileDependency}[1]{%
  \typeout{(#1)}
  \@addtofilelist{#1}
  \IfFileExists{#1}{}{\typeout{No file #1.}}
}
\makeatother

\newcommand{\ie}{i.e. }
\newcommand{\eg}{e.g.\ }

\usepackage{amsthm}
\theoremstyle{plain}
\newtheorem{theorem}{Theorem}[section]
\newtheorem{proposition}[theorem]{Proposition}
\newtheorem{lemma}[theorem]{Lemma}

\theoremstyle{definition}

\newtheorem{assumption}[theorem]{Assumption}
\theoremstyle{remark}
\newtheorem{remark}[theorem]{Remark}

\newcommand{\colorinit}[1]{{\color{Brown}#1}}

\newcommand{\colormsg}[1]{{\color{Blue}#1}}
\newcommand{\colorupd}[1]{{\color{ForestGreen}#1}}
\newcommand{\colorread}[1]{{\color{Fuchsia}#1}}

\DeclareMathAlphabet{\mathsfit}{\encodingdefault}{\sfdefault}{m}{sl}

\SetMathAlphabet{\mathsfit}{bold}{\encodingdefault}{\sfdefault}{bx}{n}

\usepackage{mathtools}

\newcommand{\sampling}{(\bx,y) \sim \pdata}

\newcommand{\fun}{f_{C,\bth}}

\newcommand{\hv}{\bh_V}

\newcommand{\xv}{\bx_V}
\newcommand{\pv}{\bp_V}

\newcommand{\mvfw}{\b{m}_{V{\text -}F}}
\newcommand{\mvbw}{\b{m}_{V{\text -}B}}
\newcommand{\xe}{\bx_E}
\newcommand{\he}{\bh_E}

\newcommand{\pe}{\bp_E}

\newcommand{\MSG}{\colormsg{MSG}}
\newcommand{\UPD}{\colorupd{UPD}}

\newcommand{\pdata}{p_{\text{d}}}

\usepackage{mathtools}

\renewcommand{\b}[1]{\mathbf{#1}}
\newcommand{\bA}{\b{A}}
\newcommand{\bB}{\b{B}}

\newcommand{\bI}{\b{I}}

\newcommand{\bP}{\b{P}}

\newcommand{\bW}{\b{W}}

\newcommand{\bb}{\b{b}}
\newcommand{\bc}{\b{c}}

\newcommand{\bh}{\b{h}}

\newcommand{\bp}{\b{p}}

\newcommand{\bx}{\b{x}}

\newcommand{\bth}{\boldsymbol{\theta}}

\newcommand{\cE}{\mathcal{E}}
\newcommand{\cF}{\mathcal{F}}
\newcommand{\cG}{\mathcal{G}}

\newcommand{\cN}{\mathcal{N}}

\newcommand{\cV}{\mathcal{V}}

\newcommand{\constr}{\mathfrak{G}} 
\newcommand{\Sth}{S_{\bth}}             %

\newcommand{\nheads}{H}                 %

\newcommand{\VP}{\mathcal{V}_P}
\newcommand{\EP}{\mathcal{E}_P}
\newcommand{\VL}{\mathcal{V}_L}
\newcommand{\EL}{\mathcal{E}_L}

\newcommand{\poolfn}{\rho}              %

\newcommand{\SthMHSA}{\Sth^{\mathrm{MHSA}}}

\newcommand{\mthd}{H-GMN }
\newcommand{\mthdT}{H-GMN}

\newcommand{\R}{\mathbb{R}}

\newcommand{\INIT}{\colorinit{INIT}}
\newcommand{\READ}{\colorread{READ}}

\newcommand{\RC}{RC}

\usepackage{cleveref}
\usepackage{algorithm}
\usepackage{algpseudocode}
\usepackage{tikz}
\usepackage{pgfplots}
\pgfplotsset{compat=1.18}
\usepackage{graphicx}
\usepackage{subcaption}
\usepackage[dvipsnames]{xcolor}

\title{Hierarchical Graph Metanetworks}

\author{
Ioannis Kalogeropoulos$^{1,3}$\\
\texttt{ioakalogero@di.uoa.gr}\\
\And
Giorgos Bouritsas$^{1,2}$\\
\texttt{g.bouritsas@athenarc.gr}\\
\And
Yannis Panagakis$^{1,2,3}$\\
\texttt{yannisp@di.uoa.gr}\\
\\
$^1$ National and Kapodistrian University of Athens\\
$^2$ Visible Machines, AI Research \& Social Awareness Center \\
$^3$ Archimedes/Athena RC, Greece\\
}

\iclrfinalcopy %
\begin{document}

\maketitle

\begin{abstract}

The emerging field of weight space learning has demonstrated the potential
to enable 
{metanetworks}, \ie models that process neural network (NN) data, 
to 
support
several steps of the machine learning lifecycle, from model optimisation to evaluation and adaptation. 
However,
this potential has yet to materialise beyond 
small-scale MLPs or CNNs: 
most current models, including message-passing based graph metanetworks (GMNs),
operate exclusively at the finest level of granularity,
processing every neuron and weight of the input model. Their cost therefore scales with the number of parameters.
 Yet NNs are \textit{inherently organised hierarchically}, with neurons forming layers and layers forming blocks and modules. Additionally, several of their properties, such as Lipschitz upper bounds and generalisation measures,
 are compositions of layer- and block-level quantities. 
To incorporate this inductive bias, we introduce Hierarchical Graph Metanetworks (H-GMNs), which leverage the input architecture's hierarchy to coarsen and refine the graph using permutation-symmetry-preserving pooling and un-pooling layers.
Further, we extend H-GMNs to Transformers by introducing a graph construction that preserves permutation symmetries and enables multi-level pooling.
Our evaluation includes new ResNet-18, ViT and varying ViT model zoos of up to 86M parameters, 
on tasks spanning model evaluation (an analytic Lipschitz bound proxy, generalisation and robustness prediction), testing (backdoor detection) and editing (pruning). Experimental results show that, compared to non-equivariant alternatives and flat GMNs, H-GMNs are markedly more sample-efficient, and their cost remains nearly constant as metanetwork depth grows, allowing deeper and more expressive processing where flat GMNs run out of memory, without compromising accuracy, which in certain cases considerably improves. Code is available at \href{https://github.com/jkalogero/hgmn}{https://github.com/jkalogero/hgmn}.

\end{abstract}

\section{Introduction}\label{sec:intro}

Every training loop of a machine learning model leaves behind a collection of artefacts -  weights, gradients, computational traces, optimisation trajectories - a considerable fraction of which are subsequently released on public model repositories. 
Notwithstanding their abundance, such artefacts have traditionally been regarded as terminal outputs of the training process and rarely as objects of analysis in their own right. It was not until recently that a growing body of work started to show that these objects, recently grouped under the term \textit{neural artefacts},
are in fact information-rich \citep{schurholt2021self, huben2024sparse}. In particular, they have been used to evaluate and edit model behaviour,
such as generalisation \citep{eilertsen2020classifying, unterthiner2020predicting}, sycophancy \citep{3692070.3692337}, hallucinations \citep{frasca2026neural} and fairness \citep{NEURIPS2020_92650b2e, karvonen2025robustly, yang2025bias}, in neural field processing \citep{10378032}, in model retrieval \citep{horwitz2025learning}, comparison \citep{pal2024model} and merging \citep{soro2026ls}.

In parallel, the structure of these objects has itself become a subject of study. Neural network (NN) parameters admit symmetries \citep{godfrey2022symmetries}, i.e. transformations that leave the function computed by the network unchanged, and they are organised according to a graph topology inherited from the underlying architecture. These observations have motivated the design of dedicated, structure-aware architectures for processing NN parameters, commonly referred to as \textit{metanetworks} (alternatively, neural functionals or meta-models) \citep{navon2023equivariant, zhou2023permutation, lim2024graph, kofinas2024graph, kalogeropoulos2024scale, pmlr-v267-vo25b, tran2025equivariant, ballerini2026weight}, predominantly within the field of Weight Space Learning (WSL), in which NN parameters constitute the primary object of study. Among these, graph metanetworks (GMNs) represent an NN as a parameter graph, whose vertices and edges correspond to neurons and weights, and process it with a graph neural network  \citep{lim2024graph, kofinas2024graph, kalogeropoulos2024scale}.

Existing GMNs, however, operate at a single level of resolution, that of individual neurons and weights, and equivariant metanetworks have, to date, been applied predominantly to modest MLPs, CNNs and Transformers. Methods that do scale to larger models, notably those that tokenise weight matrices and process the resulting sequences with a generic Transformer \citep{schurholt2024towards, soro2026ls}, account for neither the graph structure nor the symmetries of the input network, and therefore dispense with the inductive biases to which structure-aware metanetworks owe their sample efficiency. Neural networks, moreover, are constructed modularly, with channels forming layers, layers forming blocks, and blocks forming complete architectures. Accordingly, several properties of interest are naturally characterised at these coarser levels of abstraction; Lipschitz constant upper bounds, for example, arise as compositions of layer- and block-level quantities. Although hierarchical structure of this kind has been exploited in graph representation learning \citep{ying2018hierarchical,
wang2020equivariant},
it has not, to the best of our knowledge, been leveraged in the design of metanetworks.

In this paper, we propose \textit{Hierarchical Graph Metanetworks} (H-GMNs), which process the parameter graph \citep{lim2024graph, kofinas2024graph} jointly with a sequence of coarsenings derived from the architecture of the input network. \textit{Local pooling} layers consolidate each architectural component, such as a layer, an attention head or a block, into a single node, invariantly to the permutations acting within that component \citep{hecht1990algebraic} and equivariantly to permutations of interchangeable components, such as the heads of a self-attention layer. The framework is not restricted to permutations: it extends to other parameter symmetries, such as scalings \citep{godfrey2022symmetries, kalogeropoulos2024scale}, provided that all layers, including pooling and un-pooling, respect them. Since the coarsened graphs scale with the number of architectural components rather than with the parameter count, deeper and more expressive processing at the coarse levels, including graph transformers \citep{ma2023graph}, becomes computationally affordable. For editing tasks, which require outputs at the resolution of individual neurons, we introduce equivariant \textit{un-pooling} layers within a U-Net-style encoder--decoder \citep{ronneberger2015u}: un-pooling broadcasts coarse representations back to finer levels, where skip connections combine them with the fine-grained representations of the encoder. H-GMNs thereby bridge two existing paradigms and contain both as limiting cases: statistics-based metanetworks \citep{eilertsen2020classifying}, which summarise each layer through symmetric statistics before any message passing, and flat GMNs \citep{lim2024graph, kofinas2024graph}, which perform message passing exclusively on the parameter graph. Finally, to extend H-GMNs to Transformers, we propose a per-head graph construction for multi-head attention that preserves its permutation symmetries and enables pooling at the level of individual heads. Our contributions can be summarised as follows.

\begin{itemize}
    \item \textbf{A multi-resolution view of metanetworks.} H-GMNs process the parameter graph together with a sequence of architecture-derived coarsenings, and contain statistics-based metanetworks and flat GMNs as limiting cases of a single family.
    \item \textbf{Symmetry-respecting pooling and un-pooling layers}, under which predictions are invariant and edits are equivariant to neuron permutations, yielding coarse graphs whose size scales with the number of architectural components rather than with the parameter count.
    \item \textbf{A per-head parameter-graph construction for multi-head attention}, whose node groups expose both within-head and head-relabelling symmetries to pooling.
    \item \textbf{Four new model zoos} (ResNet-18, backdoored ResNet-18, ViT-S/16 and varying ViT architectures) with hyperparameter diversity, together with a task whose ground truth is known and compositional, namely a Lipschitz upper bound.
    \item \textbf{An empirical evaluation} spanning model evaluation, testing and editing, which shows that H-GMNs match or improve upon flat GMNs at lower computational cost, with gains most pronounced in low-data regimes and on the larger ViT zoos.
\end{itemize}

\section{Related Work}\label{sec:rel-work}
\noindent\textbf{Weight Space Learning \& Metanetworks.} Initial attempts to learn to process NN parameters focused on simple statistics \citep{eilertsen2020classifying, unterthiner2020predicting, martin2021predicting} or on parameter flattening \citep{schurholt2021self}. 
Concurrently, a separate line of work focused on designing intricate \textbf{Tensor-based linear layers} to account for symmetries stemming from the model architecture. \cite{navon2023equivariant, zhou2023permutation, zhou2023neural, sverdlov2025revisiting} examined \textbf{permutation} equivariance for MLPs and CNN, \cite{herrmann2024learning} 
extended this to RNNs and \cite{zhou2024universal} to general architectures,
 deriving maximally expressive equivariant layers for RNNs and Transformers. Moving beyond permutations, \cite{tran2024monomial} and \cite{pmlr-v267-vo25b} proposed \textbf{scale}-equivariant layers.
 \cite{zhang2025beyond} investigated transformer \textbf{rotation} symmetries, while more general \textbf{GL-equivariant} layers have been proposed for self-attention and LoRA weights \citep{tran2025equivariant, putterman2025gl}. Relevant works also include \citep{andreis2024set}, where 
 regularisation is used for \textbf{approximate invariance} and
 \citep{shamsian2024improved} that propose  weight-space \textbf{augmentations}.
\noindent \textbf{Graph Metanetworks (GMNs):} 
\cite{kofinas2024graph} and \cite{lim2024graph} proposed an alternative viewpoint, where NNs are converted into a special class of graphs to be processed with Graph Neural Networks (GNNs) with  symmetry breaking wherever necessary, 
focusing on MLPs, CNNs
attention, normalization, and group-equivariant linear layers. Subsequently, \cite{kalogeropoulos2024scale} adapted this paradigm to account for a variety of scaling symmetries. %
Applications of GMNs, beyond standard property prediction, include
\citep{ballerini2026weight} for NeRFs and \citep{kalogeropoulos2025metanetworks} for unsupervised model editing, while \cite{knyazev2025accelerating} used a GMN 
to forecast future weights and accelerate training.
 GMNs are adaptable to new architectures, take into account the underlying NN graph topology and have demonstrated better experimental performance. Therefore, they are selected as a backbone for the present work.

\noindent\textbf{Towards larger models.} However, none of the above methods has been applied to networks beyond the tens-of-thousands-of-parameters regime. The sole exception is \citet{knyazev2025accelerating}, where the largest model used for meta-training was a GPT2-style Transformer ($\sim$1.6M parameters) and the largest model evaluated on was a Llama3-style model ($\sim$111M parameters), for which $5\%$ edge sampling was necessary and inference took over $10$ sec/prediction. Towards this direction,  \cite{schurholt2024towards}  proposed an autoencoder, named \textit{SANE}, to learn self-supervised representations of tokenised weight matrices, a work extended in \citep{falk2025learning}, and adapted for learnable model merging \citep{soro2026ls} and dataset-NN parameter representation alignment \citep{asefaw2026weightclip}. Nevertheless, these methods lack appropriate inductive biases (graph-structure aware, equivariant) and typically need larger training datasets and tend to underperform (see \cref{sec:exp}). Finally, probing methods sidestep parameter count, representing a model via its responses to learned probes, fed to the whole network \citep{kahana2025deep} or a single layer \citep{horwitz2025learning}.

\noindent\textbf{Hierarchical GNNs.} Flat message-passing GNNs suffer from over-squashing and a limited effective receptive field \citep{alon2020bottleneck,topping2021understanding,di2023over}.
A common remedy is to process the graph at multiple scales; hierarchical methods differ mainly in the origin of the hierarchy.
\textbf{Learnable coarsening algorithms.} Dense soft-clustering, \eg DiffPool
\citep{ying2018hierarchical} and MinCutPool \citep{bianchi2020spectral}, and sparse
node selection, \eg graph U-Nets \citep{gao2019graph}, SAGPool \citep{lee2019self}, learn the coarsening end-to-end. 
\textbf{Non-learnable coarsening algorithms.} Another line computes the hierarchy from the graph
topology with generic coarsening algorithms, \eg 
Graclus \citep{defferrard2016convolutional},
Louvain \citep{rampavsek2021hierarchical, zhong2023hierarchical}, 
BFS-based pooling \citep{cao2023efficient}
and recursive clustering \citep{vonessen2024next, finder2025improving}. These are cheap and stable but agnostic to the data distribution. Scalability-oriented partitioning
\citep{chiang2019cluster,liao2018graph} and virtual-node rewiring
\citep{qian2024probabilistic} are related but do not target hierarchical representation learning.
\noindent\textbf{Hierarchies provided by the data.} When the data has an intrinsic multi-level organisation, it can be used directly: 
junction trees \citep{fey2020hierarchical}, protein
residue-backbone-atom levels \citep{wang2022learning}, 
or given subgraphs \citep{kim2024translating}.  
\citet{wang2020equivariant} formally characterises the symmetries of such nested structures. This is the strategy we also adopt, given the pre-defined modular manner, in which NNs are designed.

\section{Preliminaries: Graph Metanetworks}\label{sec:prelims}
\label{sec:pg-lim}

Let $f_{C, \bm{\theta}}(\cdot)$ be an NN parametrised by its computational graph $C$ and a vector of learnable parameters $\bm{\theta}$, with $f_{C, \bm{\theta}} \in \cF$, where $\cF$ is a function space. The bigger picture surrounding our approach is the design of \textit{metanetworks},  learnable functions that process NNs, \ie  $\mathfrak{F}: \cF \to \cF'$, where $\cF'$ can be a vector space or a function space itself, depending on the task.

\noindent\textbf{Graph construction.} To model $\mathfrak{F}$, we follow a common WSL paradigm that treats NN datapoints as \textit{graphs} \citep{lim2024graph, kofinas2024graph, kalogeropoulos2024scale}. Therefore, the first step is to construct a graph faithfully representing each NN by a procedure 
$\mathfrak{G}: \cF \to \cG$, where $\cG$ is a graph space,
that receives an NN $f_{C, \bm{\theta}}$ and yields a \textit{parameter graph} $\mathfrak{G}(f_{C, \bm{\theta}}) = G = (\cV, \cE, \xv, \xe)$, carefully designed such that $\fun$ can be retrieved.
Typically, $\constr$ traverses the computational graph $C$ layer-by-layer, populating the set $\cV$ with one node for each output channel of each layer and the set $\cE$ with one edge for each pair of input-output channels that interact (\ie when the latter is a function of the former). The exact wiring depends on the layer type: \textit{channel-mixing} layers (\eg linear, convolution, attention projections), 
connect input with output nodes by a complete bipartite graph, and
\textit{channel-wise} layers (\eg BatchNorm, LayerNorm), 
connect them node-wise. 
Regarding the features, although $\mathfrak{G}$ differs between architectures, a general rule of thumb is that biases and weights become vertex and edge features, respectively. 
Additionally, the construction often allows 
$\xv, \xe$ to inherit the symmetries from the NN's parameters (\eg permutations, scalings). 
Regardless of the input $\fun$, the graph $G$ exhibits a structural regularity: it is, by construction, a directed (edges start from input channel nodes and terminate at output ones), acyclic (no node points back to a node corresponding to a channel of a previous NN layer), multi-partite graph (nodes corresponding to channels of the same layer are never connected). \Cref{fig:PG} depicts the Parameter Graph for a ResNet18 block.
A detailed description of the graph construction can be found in \cref{app:graph-constr}.

\noindent\textbf{Message Passing.} The second step is to sequentially process the parameter graph with multiple message passing layers (Graph Metanetworks - GMNs). These closely follow the design principles of GNNs consisting of vertex- and edge- message and update functions, as well as positional encodings for symmetry breaking, all potentially modified to account for specific NN symmetries. Please refer to \citep{lim2024graph, kalogeropoulos2024scale} for additional details. With a slight notation abuse, we will denote a GNN with message passing layers also with $\mathfrak{F}: \cG \to \cG$, with  $\mathfrak{F}(\cV, \cE, \bh_V, \bh_E) = (\cV, \cE, \bh_V', \bh_E')$, where $(\cV, \cE)$ the graph topology that remains unchanged and $\bh_V, \bh_E$, $\bh_V', \bh_E'$, the input and output vertex and edge features. The last step is a \textbf{Readout} function $\mathfrak{R}: \cG \to \cF'$ that projects the output of the last metanetwork layer to the output space of interest. This will be a prediction head combined with either a global invariant aggregation for NN behaviour prediction, or with the inverse function $\mathfrak{G}^{-1}$ to project back to an output NN, for NN editing.

\section{Hierarchical Graph Metanetworks}\label{sec:method}

NNs are compositional: channels form layers, layers form blocks, and blocks form entire modules. To account for this, we propose a progressive coarsening scheme 
producing a sequence of graphs $(G_i)_{i=1}^M$, where $M$ is the number of levels. Node and edge features of graphs higher in the hierarchy consolidate information from lower levels by learnable \textit{graph pooling layers} and, in certain cases, can pass on information to lower levels by \textit{graph unpooling}.
Unlike many hierarchical models, which additionally learn how to coarsen the graph 
\citep{ying2018hierarchical, lee2019self}, our graph sequence is derived from the input's architecture: each step aggregates the nodes and edges of the same architectural component (e.g.\ a layer, attention head, or block). Besides giving the model direct access to component-level structure, coarsening yields graphs that scale with the NN depth, instead of $\text{width}^2*\text{depth}$, allowing for deeper processing using wider GNN layers or even Graph Transformers \citep{ma2023graph} and extending the receptive field to capture longer-range interactions.

\noindent\textbf{Pooling Layer.} Let $G_i = (\cV_i, \cE_i, \bh_V^{(i)}, \bh_E^{(i)})$ be a graph and $(\cV_{i+1}, \cE_{i+1})$ a coarsened topology. A \emph{pooling} layer 
$\mathrm{Pool}_i:\cG \to \cG$, with  $\mathrm{Pool}_i(\cV_i, \cE_i, \bh_V^{(i)}, \bh_E^{(i)}) = (\cV_{i+1}, \cE_{i+1}, \bh_V^{(i+1)}, \bh_E^{(i+1)})$
is specified by a surjective node-grouping map $g_i : \cV_i \to \cV_{i+1}$ and node and edge pooling functions $\poolfn_{V,i}, \poolfn_{E,i}$, each aggregating the embeddings of a group to a single embedding:
\begin{align}\label{eq:pooling}
  \bh_V^{(i+1)}(u)     
  = \poolfn_{V,i}\left( \left(\bh_V^{(i)}(v)\right)_{g_i(v) = u} \right), 
  \bh_E^{(i+1)}(u,w) 
  = \poolfn_{E,i}\left( \left(\bh_E^{(i)}(s,t)\right)_{(g_i(s), g_i(t)) = (u,w)} \right), 
\end{align}
with $u \in \cV_{i+1}, (u,w) \in \cE_{i+1}$ and
${\cE_{i+1} = \{ (u,w) : \exists s,t \in \cV, \text{ with } (s,t)  \in \cE_i,\ g_i(s), g_i(t) = u, w \}}$.

Each $g_i$ is derived from the architecture and depends on the hierarchy levels it connects. A simple \mthd\ allows communication only between consecutive layers (akin to hierarchical CNNs) and composes pooling  with $i$-level GNNs $\mathfrak{F}_i: \cG \to \cG$ acting on $G_i$, $i = 1, \dots, M$, where $G_1 = G$:
\begin{equation}\label{eq:encoder}
  \mathfrak{F}_{\text{pred}}(\fun) = \mathfrak{R} \circ \mathfrak{F}_M \circ \mathrm{Pool}_{M-1} \circ \dots \circ \mathrm{Pool}_1 \circ \mathfrak{F}_1 \circ \mathfrak{G}(\fun),
\end{equation}

Each $\mathfrak{F}_i$ comprises $k_i \geq 0$ message-passing layers, with $k_i = 0$ corresponding to the identity map.

\noindent\textbf{Resolution as a design axis.} The hierarchy exposes a choice that existing metanetworks fix implicitly, namely the resolution at which processing takes place. H-GMNs are indexed by the number of levels $M$ and the number of message-passing layers $k_i$ applied at each level. With $k_i = 0$ at every level, fixed symmetric statistics as $\rho_{V,1}, \rho_{E,1}$, a final pooling that concatenates all layer embeddings into a single node, and an MLP readout, an H-GMN summarises every layer before any message passing and recovers StatNN \citep{unterthiner2020predicting}, up to the choice of statistics and predictor. With a single level ($M = 1$), it reduces to a flat GMN. Intermediate choices spend a few layers at neuron resolution, where interactions between individual weights are visible, and the remaining layers at coarser resolutions, where the graph is orders of magnitude smaller (see \cref{tab:graph_stats}).

\noindent\textbf{Unpooling Layer.}
Consider given vertex and edge sets $\cV_i, \cE_i$, a coarsening map $g_i : \cV_i \to \cV_{i+1}$, 
and embeddings $\bh_V^{(i+1)}, \bh_E^{(i+1)}$ of $G_{i+1}$. The \emph{unpooling} layer $\mathrm{UnPool}_i: \cG \to \cG$, with $\mathrm{UnPool}_i(\cV_{+1}, \cE_{i+1}, \bh_V^{(i+1)}, \bh_E^{(i+1)}) = (\cV_{i}, \cE_{i}, \bh_V^{(i)}, \bh_E^{(i)})$  copies the embedding of a node in $G_{i+1}$ to every node and edge that maps to it based on $g_i$:
\begin{align} \label{eq:unpooling}
  \bh_V^{(i)}(v)     = \bh_V^{(i+1)}\left(g_i(v)\right), \  v \in \cV_i, &&
  \bh_E^{(i)}(s,t) = \bh_E^{(i+1)}\left((g_i(s), g_i(t))\right), && (s,t) \in \cE_i.
\end{align}

\paragraph{Editing pipeline.} 
Editing tasks require output at the same resolution as the input. Naively broadcasting an intermediate level representation back to the parameter graph would assign every neuron in a layer the same coarse embedding, discarding the fine-grained distinctions an edit might require. Instead, we employ a U-Net-style encoder-decoder \citep{ronneberger2015u, gao2019graph}: the method in \cref{eq:encoder} serves as an encoder, and a decoder climbs up the resolution levels, 
via unpooling, concatenating with the encoder's embeddings and further processing at each resolution,
giving the final GNN stage access to both global structure and local embeddings.

In particular,
the encoder is a H-GMN as in \cref{eq:encoder} without readout, and $G_{i} = (\cV_i, \cE_i,\bh^{(i)})$ is the GNN output at level $i$, where we use the shorthand
$\bh^{(i)} = (\bh_V^{(i)}, \bh_E^{(i)})$  for the encoder embeddings. The decoder returns to the finest level, unpooling at each step, combining with the encoder embeddings of that level through a skip connection and processing with a GNN layer $\tilde{\mathfrak{F}}_i$ on $\tilde{G}_i = (\cV_i, \cE_i,\tilde\bh^{(i)})$, where $\tilde{\bh}^{(i)} = (\tilde{\bh}_V^{(i)}, \tilde{\bh}_E^{(i)})$ are the decoder embeddings:
 \begin{equation}\label{eq:decode}
 \begin{split}
  &\bh^{(1)} = \mathfrak{F}_1(\mathfrak{G}(\fun)), \qquad \bh^{(i)} = \mathfrak{F}_i\big(\mathrm{Pool}_{i-1}(G_{i-1})), \quad i = 2, \dots, M,  \quad \tilde\bh^{(M)} = \bh^{(M)}, \\
  &
  (\cV_{i}, \cE_{i}, \tilde\bh_{a}^{(i)}) = \mathrm{UnPool}_{i}(\tilde{G}_{i+1}), 
  \quad
  \tilde\bh_b^{(i)} = \tilde\bh_a^{(i)} \,\|\, \bh^{(i)},
   \quad
   \tilde\bh^{(i)} = \tilde{\mathfrak{F}}_i\big(\cV_i, \cE_i,\tilde{\bh}_b^{(i)}\big),
\end{split}
\end{equation}
where $i =1, \dots, M-1$ and concatenation $\|$ happens across the channel dimension.
The output of the pipeline is $\mathfrak{F}_{\mathrm{edit}}(\fun) = \constr^{-1}\big(\psi(\tilde\bh^{(1)})\big)$, where $\psi$ is a task-specific node- and edge-wise head.

\section{ Scaling to Transformers: Multi-Head Self-Attention Graphs}\label{sec:graph-con}

Multi-head self-attention layers (MHSA) are understudied within the GMN paradigm.
\cite{lim2024graph} and \cite{kofinas2024graph}  treat attention projection as standard linear layer parameters, imperfectly modelling permutation symmetries. \cite{knyazev2025accelerating} rectifies this and is closer to our construction, but introduces unnecessary auxiliary vertices and edges.
Below, we provide a general graph construction method that seamlessly allows MHSA processing by H-GMN.

\noindent\textbf{Self-attention. } Consider an input containing $n$ elements $\bx \in \mathbb{R}^{d_{\text{in}} \times n}$ and a self-attention function $f_{C, \bth}(\bx) =  (\bW_v \bx) \sigma\big((\bW_q \bx)^T (\bW_k\bx)\big) $, where $\bW_q, \bW_k \in \mathbb{R}^{d_k \times d_{\text{in}}}$ project queries and keys into the same space to compute the scores (dot products), $\sigma(\cdot)$ a non-linear function converting scores to attention coefficients (\eg softmax after scaling with $1/\sqrt{d}$) and $\bW_v \in \mathbb{R}^{d_{\text{v}} \times d_{\text{in}}}$ the value matrix (biases are omitted for simplicity). To construct the parameter graph, we follow the standard linear layer strategy, \ie vertex-channel/edge-weight mapping (similar to CNNs \citep{lim2024graph}, the dimension $n$ should not affect the graph, since it is data-dependent), albeit with crucial differences. 

Let $\cV$ be the input vertices corresponding to $\bx$, \ie $|\cV| = d_{\text{in}}$. First, we construct a set of vertices $\cV_v$, with 
$|\cV_v| = d_v$ and connections $\cE_v = \cV \times \cV_v$ with $\bx_E(i,j) = [\bW_v(j,i)]$ as with all linear layers.
Then, we construct a set $\cV_k$, with 
$|\cV_k| = d_k$ and connections $\cE_k = \cV \times \cV_k$ with corresponding edge features $\bx_E(i,j) = [\bW_q(j,i), \bW_k(j,i)]$.
This reflects that these  weight matrices are synchronised 
and share permutations. Finally, we create connections $\cE_{kv} = \cV_k \times \cV_v$ with constant features and amend the latter two edge sets with different edge types to discern them from linear layers.

\begin{figure}[t]
    \centering
    \includegraphics[trim=0 60pt 0 60pt, clip, width=0.7\textwidth]{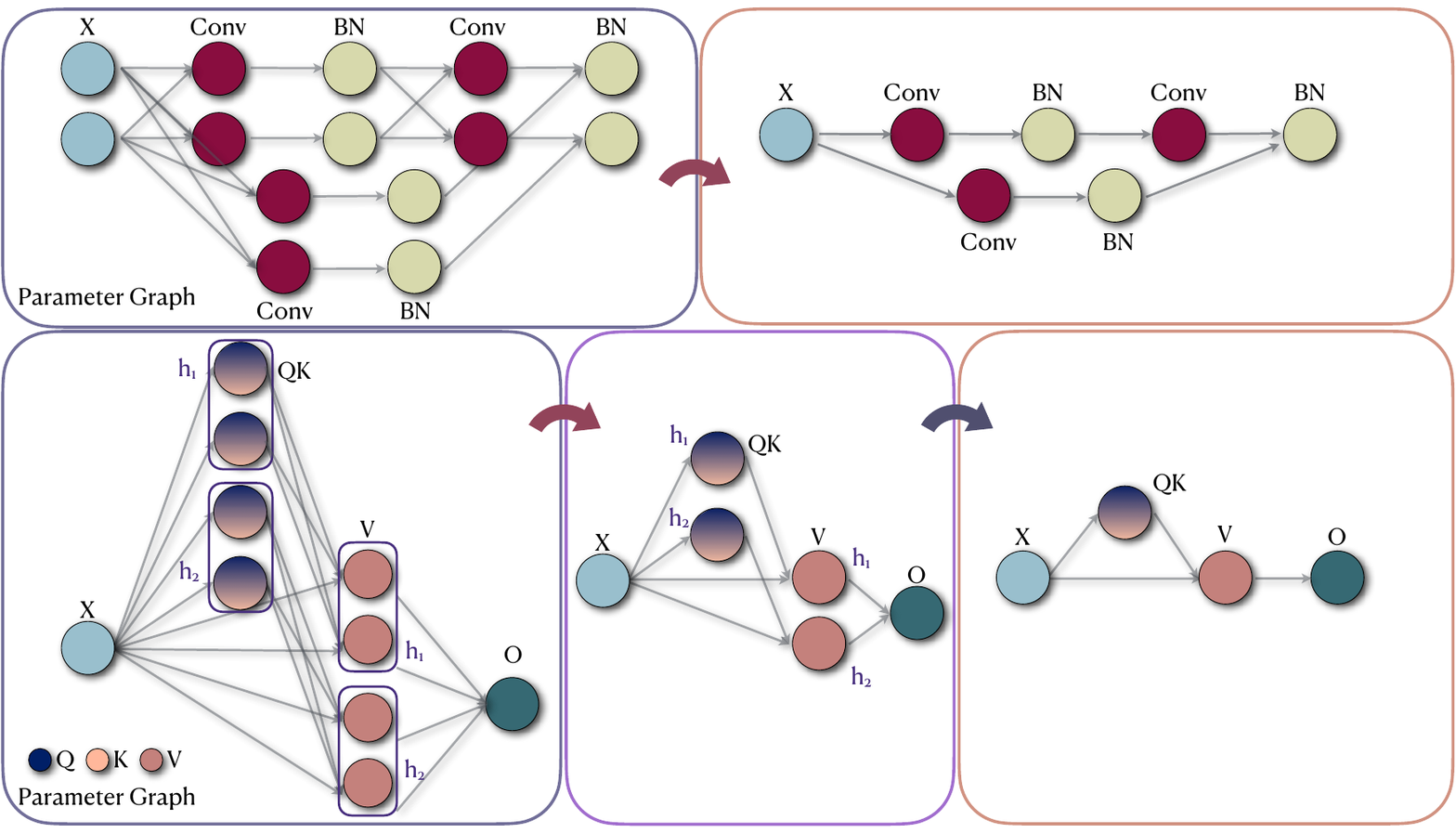}
    \caption{\textbf{Top}: Pooling  on ResNet basic block. Red arrow indicates pooling across the width of each layer, resulting in the coarsened layer graph. \textbf{Bottom}: Pooling on an MHSA block. Pooling across the width leaves head vertex sets visible. Further pooling across heads leads to the layer graph level.}
    \label{fig:PG}
\end{figure}

\noindent\textbf{Multi-head. } In the classical multi-head setting, all head outputs are further projected and mixed into a single output, \ie $f_{C, \bth}(\bx) =  \sum_{h}\bW^h_o(\bW^h_v \bx) \sigma\big((\bW^h_q \bx)^T (\bW^h_k\bx)\big) $, where $\bW^h_o \in \mathbb{R}^{d_{\text{out}} \times d_v}$.
Therefore, we repeat the creation of new vertices and edges for each head separately and additionally create a new vertex set $\cV_o$, with $|\cV_o| = d_{\text{out}}$, with connections $\cE_o = \{(i,j) | i \in \cV^h_v, j \in \cV_o, \forall h\}$, with edge features $\bx_E(i,j) = \bW^h_o(j,i), $where the superscript $h$ has been used to extend the above notation (see \cref{fig:PG} for an illustration). 
In \cref{sec:MHSA_perm}, we provide a proof and additional explanations on how parameter symmetries are inherited to the vertex/edge features for equivariant processing.

\section{Analysis \& Instantiations}\label{sec:symmetries}

\noindent\textbf{Pooling and Unpooling Symmetries.} 
Instantiating pooling layers in a symmetry-respecting manner is \underline{tightly related to how vertices are grouped at each level of hierarchy}.  To simplify, we focus on \textit{permutation symmetries}.
The most common case is when the grouping contains \textit{all} vertices that are equivalent to each other, \eg 
coarsening the parameter graph into a layer graph. To make the metanetwork invariant, the pooling function should be invariant to valid permutations, and those permutations are no longer applicable in the next levels of hierarchy (\textit{collapsed symmetries}). Another case arises when the grouping contains \textit{a subset} of equivalent vertices, \eg when separately grouping the vertices corresponding to different attention heads. Again here, invariance to valid permutations is necessary, but the resulting vertices remain equivalent to the ones that did not participate in the grouping. In other words, certain permutations carry over in the next level of hierarchy, and subsequent layers may be equivariant to them (\textit{surviving symmetries}). To achieve metanetwork equivariance when editing, the concatenation trick of \cref{eq:decode} is key, as the original symmetries are carried over to the output 
and remain unaffected when broadcasting invariants with unpooling.

In practice, we use classical permutation-invariant aggregators for all pooling functions and account for non-valid permutations (when pooling inputs that are not equivalent, \eg pooling vertices of the layer graph) with positional encodings. For both vertex and edge pooling 
we found that a simple DeepSets architecture 
\citep{zaheer2017deep}
with several symmetric statistics (e.g.\ mean, max, standard deviation as in PNA \citep{corso2020principal}) was sufficient for our experiments. Note that our approach can be seamlessly extended to continuous symmetries, such as scaling, by modifying the pooling functions to be invariant to those as well, \eg see \citep{kalogeropoulos2024scale}.

\textbf{ResNet Hierarchies.} The simplest case of an MLP or a CNN admits a two-level hierarchy, from parameters to layers. Additional interest is found in ResNets, where an extra level arises due to their block structure (\Cref{fig:PG} illustrates a ResNet-18 basic block). $\mathrm{Pool}_1$ groups neurons
collapsing each layer into one vertex and each weight matrix into one edge; then   $\mathrm{Pool}_2$ 
groups layers of a residual block into one vertex and one edge. Since $\mathrm{Pool}_1$ already collapses every symmetry of $\bm\theta$, the pipeline is already invariant at this point and subsequent functions can be chosen arbitrarily.

\textbf{Transformer hierarchies.} NNs that contain MHSA blocks add an extra hierarchy level, that of heads. $\mathrm{Pool}_1$ groups neurons at each block and head, collapsing within-head symmetries, but retaining cross-head ones, as mentioned before. Thus, subsequent metanetwork layers are equivariant to head permutations, until 
 $\mathrm{Pool}_2$ that groups heads into a layer graph, as above (\cref{fig:PG}, bottom).

\noindent\textbf{Complexity.}\label{sec:complexity} To compare the complexity of a flat GMN with H-GMN, consider
an input MLP with $(L,d)$ depth/width processed by flat GMN of $(K,w)$ depth/width vs a 2-level H-GMN with $(k_1, w_1, k_2, w_2)$ depth/width per layer and a pooling operation.
Then,  $G$ will have $(|\cV|, |\cE|) = (L d, L d^2)$ vertices and edges. First, the flat GMN scales linearly with the number of edges, and therefore will have $O(L d^2 Kw)$ memory and $O(L d^2 Kw^2)$ time complexity. 
In contrast, the coarsened graph has 
 $L$ vertices and edges and the total memory complexity is $
 O(Ld^2(k_1+1)w_1 + Lk_2w_2)$ and the time complexity is $
 O(Ld^2(k_1+1)w_1^2 + Lk_2w_2^2)$.
The quadratic dependence on $d$ multiplies only $k_1$, whereas the second term is independent of $d$. Since large networks are usually wide, keeping $k_1$ small leads to significant savings compared to a flat $K$-layer GMN.
Note also that
communication across remote nodes is costly for the flat GMN, which will require $K\geq L$, contrary to H-GMN, where communication can happen at the second level and the number of layers can be reduced depending on the hierarchy; in the worst case, $k_2 \geq L$, which incurs a much smaller cost (empirical time and memory measurements in \cref{app:time-mem}).

\section{Experiments}\label{sec:exp}
\noindent\textbf{Datasets.}

We evaluate \mthd on model zoos of ResNet18 \citep{he2016deep} and ViT \citep{dosovitskiy2020image} models. Although both architectures appear in prior zoos \citep{schurholt2024towards,falk2025model}, these vary only the random seed and are not fully publicly available, so we construct our own by also varying the learning rate and weight decay. ResNet18 models are trained on CIFAR10 and ViT models on CIFAR100 \citep{Krizhevsky2009LearningML}. We also construct a dataset of varying ViT models, with parameters ranging in $[\sim2\text{M}, \sim86\text{M}]$. Construction details are in \Cref{app:model-zoo-construction}, and graph statistics are in \Cref{tab:graph_stats}. All datasets will be publicly released.

\begin{table}[h!]
\centering
\caption{Fixed zoos graph size in different hierarchy levels. Number of checkpoints: ResNet18 8,000; MNIST ViT 15,689; CIFAR100 ViT 3,640.}
\label{tab:graph_stats}
\small
\setlength{\tabcolsep}{4pt}
\resizebox{\textwidth}{!}{%
\begin{tabular}{@{}l*{3}{ccc}@{}}
\toprule
 & \multicolumn{3}{c}{ResNet18 ($\sim$11M)} & \multicolumn{3}{c}{MNIST ViT ($\sim$20k)} & \multicolumn{3}{c}{CIFAR100 ViT ($\sim$21M)} \\
\cmidrule(lr){2-4} \cmidrule(lr){5-7} \cmidrule(lr){8-10}
Graph Type & $|V|$ & $|E|$ & Diameter & $|V|$ & $|E|$ & Diameter & $|V|$ & $|E|$ & Diameter \\
\midrule
Parameter Graph & $10,125$ & $1,405,184$ & $18$ & $619$ & $16,896$ & $16$ & $46,951$ & $19,817,472$ & $27$ \\
Head Graph      & --     & --        & --   & $23$ & $28$ & $16$ & $208$    & $363$        & $27$ \\
Layer Graph     & $43$     & $50$        & $18$ & $19$ & $20$ & $16$ & $88$     & $123$        & $27$ \\
ResBlock Graph     & $10$     & $9$        & $9$ & -- & -- & -- & --     & --        & -- \\
\bottomrule
\end{tabular}
}
\end{table}

\noindent\textbf{Lipschitz constant upper-bound proxy prediction.}

To explicitly evaluate the importance of hierarchy as an inductive bias, we designed an experiment around a task that fundamentally demands a hierarchical understanding. Specifically, we train a metanetwork to 
predict a known ground-truth function: a spectral-norm proxy of the network's Lipschitz constant upper bound, computed compositionally from layer- and block-level quantities \cref{eq:lipschitz}.

\begin{figure}[h]
    \centering
    \begin{minipage}[]{0.495\textwidth}
        \centering
    \includegraphics[width=\linewidth]{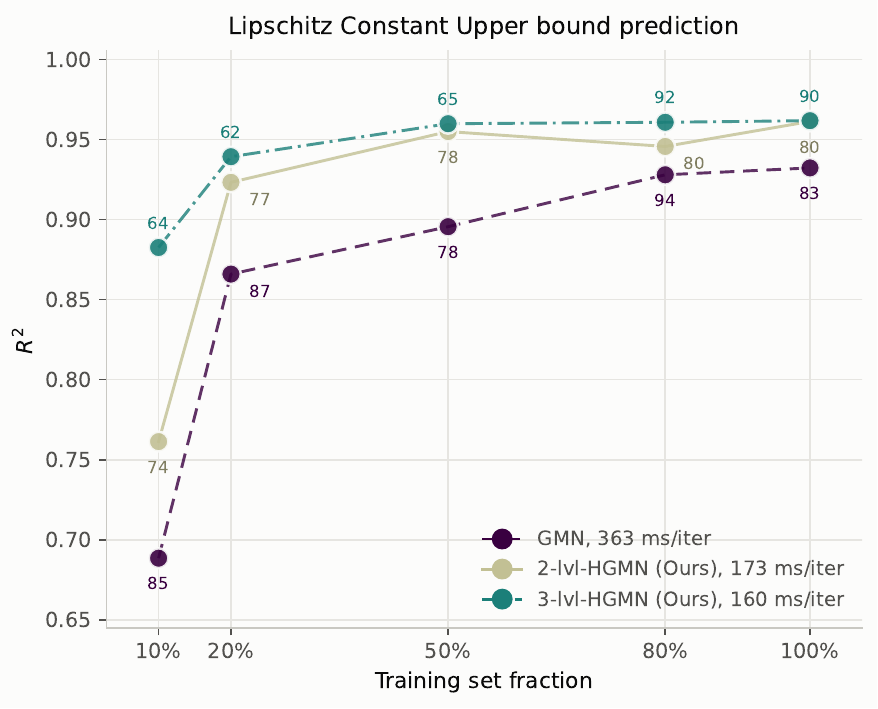}
        \caption{Coefficient of determination ($R^2 \uparrow$) on the test split for different training set sizes.}
        \label{fig:lip-pred}
    \end{minipage}\hfill
    \begin{minipage}[]{0.495\textwidth}
        \centering
    \includegraphics[width=\linewidth]{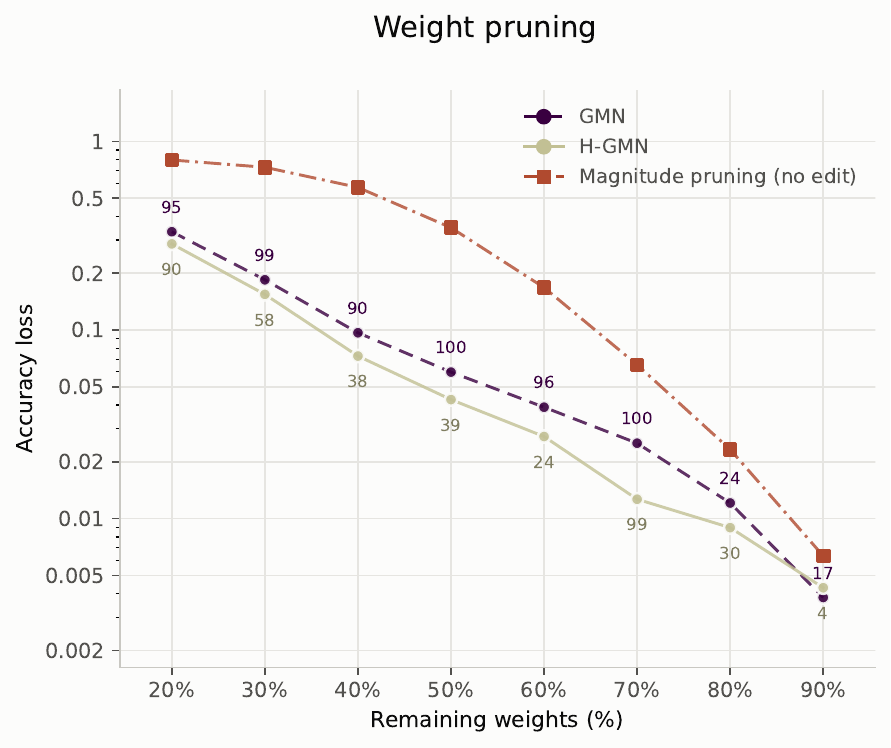}
        \caption{Average test accuracy drop (measured as $\frac{\text{acc}_{\bth} - \text{acc}_{\tilde{\bth}}}{\text{acc}_{\bth}}$) across pruning ratios. 
        }
        \label{fig:pruning}
    \end{minipage}
\end{figure}

We compare, across different training set sizes, the performance of plain GMN with \mthd\  operating on different levels of hierarchy, keeping all models under the same parameter budget,
obtained by using a total of 3 GNN layers in each case.
As shown in \cref{fig:lip-pred}, 
3-lvl-H-GMN
operating on 3 hierarchy levels (parameter, layer, and ResNet block),
\textbf{achieves comparable performance with the flat 
GMN
using only one-fifth of the training data, with $2\times $ faster inference time}. 
2-lvl-H-GMN (parameter, layer)
performs consistently better than 
GMN
with both models demonstrating improved performance as the training set size increases. \textbf{H-GMNs also converge faster} - see depictions
next to each point (earliest epoch after which performance gains remain below a tolerance~$\epsilon$).

\noindent\textbf{Generalisation prediction.}
A foundational task in WSL, generalisation prediction seeks to estimate a model's test accuracy directly from its parameters. 
We also report the \mthdT -\RC{} variants, which instantiate $\mathfrak{F}_1$ with a row--column layer adapted from \citet{hartford2018deep} rather than a GMN layer: each edge is updated from the mean embeddings of its row, its column and its whole weight matrix, at a lower cost than message passing (Appendix~\ref{app:rc}).
Beyond the inductive bias of \mthd, this experiment clearly depicts the improved scalability of our method, as we evaluate on both ResNet18 ($\sim$11M) and ViT ($\sim$21M) models.
\textbf{ResNet18}: \mthd matches $\mathrm{GMN}$ when the full training set is available, and both clearly outperform StatNN and SANE.
The advantage of H-GMN emerges as training data becomes scarce: the $\tau$ gap over GMN grows from $0.006$ at 50\% of the training checkpoints to $0.116$ at 10\%. 
Notably, even when flat 
GMN
suffices, collapsing the parameter graph into an orders of magnitude smaller graph  (see \Cref{tab:graph_stats}) \textit{incurs no performance degradation}, contrary to what one might expect. This enables more memory- and time-efficient processing, allowing the method to scale to much larger graphs.
The poor performance of SANE is attributed mainly to the $\sim 5$ times smaller training dataset than the one used by \cite{schurholt2024towards}.
\textbf{ViT:} \textbf{Our method achieves the best $\tau$ and $R^2$ on both zoos, and its margin grows with model size.}
On MNIST ViT, introduced with 
Transformer-NFN
\citep{tran2025equivariant}, 3-lvl-\mthd improves $\tau$ over that method by $0.013$. Employing 
Transformer-NFN
on the $\sim21$M-parameter CIFAR100 ViT becomes intractable.
On this dataset \textbf{3-lvl-\mthd significantly outperforms all other methods by a large margin}. Crucially, due to the size of the CIFAR100 ViT, increasing the number of GNN layers was prohibitive in terms of memory and runtime. In fact, a model with more than one GNN layer did not fit in a single 64\,GB NVIDIA H100 GPU, even with automatic mixed precision and minimal batch size and hidden dimension. On the coarsened graphs, by contrast, using deeper and wider GNNs, or even Graph Transformers, introduces negligible computational overhead. Using a single GNN layer at the fine-grained resolution and reserving more expressive, but costlier, layers for the coarser ones is precisely what makes our method more scalable than existing symmetry-aware alternatives. The \RC{} variants make this single layer cheaper: on all fixed zoos they perform on par with their GMN counterparts (within $0.01$ in $\tau$), while using about $60\%$ less GPU memory and running up to $2.2\times$ faster (\Cref{app:time-mem}). We also apply GRIT \citep{ma2023graph} directly on the Parameter Graph (GMN (GRIT)). Finally, we evaluate on the Varying ViT zoo, using \mthd-\RC{} for its lower memory footprint (\Cref{app:time-mem}). Results in \Cref{tab:gen_pred} regard training on a subset of models with parameters in $[\sim2\text{M}, \sim48\text{M}]$, since the GMN baseline on larger models required much greater computational resources for a complete training step. The final evaluation was tested on unseen architectures in the same parameter range. Evidently, \mthd outperforms all baselines.

\begin{table}[h]
\centering
\caption{Generalisation prediction.  We report Kendall's rank correlation ($\tau \uparrow$),  coefficient of determination ($R^2 \uparrow$), and convergence epoch (Ep.).
``--'': architecturally inapplicable baseline.}
\label{tab:gen_pred}
\resizebox{\textwidth}{!}{%
\begin{tabular}{@{}l*{6}{ccc}@{}}
\toprule
Method &
\multicolumn{3}{c}{\begin{tabular}[c]{@{}c@{}}ResNet18-0.1\end{tabular}} &
\multicolumn{3}{c}{\begin{tabular}[c]{@{}c@{}}ResNet18-0.5\end{tabular}} &
\multicolumn{3}{c}{\begin{tabular}[c]{@{}c@{}}ResNet18\end{tabular}} &
\multicolumn{3}{c}{\begin{tabular}[c]{@{}c@{}}MNIST ViT\end{tabular}} &
\multicolumn{3}{c}{\begin{tabular}[c]{@{}c@{}}CIFAR100 ViT\end{tabular}} &
\multicolumn{3}{c}{\begin{tabular}[c]{@{}c@{}}Varying ViT\end{tabular}} \\
\cmidrule(lr){2-4} \cmidrule(lr){5-7} \cmidrule(lr){8-10} \cmidrule(lr){11-13} \cmidrule(lr){14-16} \cmidrule(lr){17-19}
 & $\tau$ & $R^2$ & Ep. & $\tau$ & $R^2$ & Ep. & $\tau$ & $R^2$ & Ep. & $\tau$ & $R^2$ & Ep. & $\tau$ & $R^2$ & Ep. & $\tau$ & $R^2$ & Ep. \\
\midrule
StatNN 
& $0.677$ & $0.684$ & $92$ & $0.824$ & $0.884$ & $78$ & $0.837$ & $0.890$ & $82$ & $0.879$ & $0.989$ & $181$ & $0.877$ & $0.811$ & $94$ & $0.795$ & $0.554$ & $\mathbf{77}$ \\
Transformer-NFN 
& -- & -- & -- & -- & -- & -- & -- & -- & -- & $0.906$ & $0.987$ & $\mathbf{102}$ & -- & -- & -- & -- & -- & -- \\
GMN & $0.744$ & $0.840$ & $93$ & $0.886$ & $\mathbf{0.943}$ & $94$ & $0.900$ & $\mathbf{0.953}$ & $97$ & $0.916$ & $0.996$ & $190$ & $0.804$ & $0.673$ & $\mathbf{84}$ & $0.577$ & $0.491$ & $85$ \\
GMN (GRIT) 
& $0.764$ & $0.862$ & $96$ & $0.874$ & $0.924$ & $82$ & $0.889$ & $0.942$ & $91$ & $0.863$ & $0.996$ & $105$ & -- & -- & -- & -- & -- & -- \\
\midrule
SANE
& $0.584$ & $0.477$ & - & $0.614$ & $0.566$ & -- & $0.633$ & $0.604$ & -- & -- & -- & -- & -- & -- & -- & -- & -- & -- \\
\midrule
\midrule
2-lvl-\mthdT (Ours) & $0.846$ & $0.927$ & $80$ & $\mathbf{0.892}$ & $0.941$ & $74$ & $\mathbf{0.902}$ & $0.951$ & $82$  & $0.917$ & $0.997$ & $196$ & $0.862$ & $0.872$ & $97$ & -- & -- & -- \\
2-lvl-\mthdT-\RC (Ours) & $0.837$ & $0.900$ & $81$ & $0.886$ & $0.938$ & $86$ & $0.900$ & $0.952$ & $85$ & $0.919$  & $0.997$ & $184$ & $\mathbf{0.940}$ & $0.973$ & $99$ & $\mathbf{0.863}$ & $\mathbf{0.962}$ & $78$ \\
3-lvl-\mthdT (Ours) & $0.850$ & $0.906$ & $\mathbf{61}$ & $0.881$ & $0.927$ & $\mathbf{63}$ & $0.893$ & $0.945$ & $\mathbf{50}$ & $\mathbf{0.919}$ & $\mathbf{0.998}$ & $178$ & $0.927$ & $\mathbf{0.979}$ & $86$ & -- & -- & -- \\
3-lvl-\mthdT-\RC (Ours) & $\mathbf{0.860}$ & $\mathbf{0.928}$ & $75$ & $0.882$ & $0.939$ & $65$ & $0.899$ & $0.942$ & $56$ & $0.916$ & $0.997$ & $193$ & $0.929$ & $0.948$ & $96$ & $0.849$ & $0.941$ & $86$ \\
\bottomrule
\end{tabular}%
}
\end{table}

\noindent\textbf{Robustness prediction.}
Beyond accuracy, a meaningful task for metanetworks is predicting robustness.
We compute mean Corruption Error (mCE), following \cite{hendrycks2019benchmarking}, and treat mCE as the ground-truth labels to train our metanetwork. Implementation details can be found in \Cref{app:exp-details-mce}. 
As shown in \Cref{tab:mce}, 3-lvl-\mthd outperforms GMN on ResNet18 in the low-data regime and converges consistently faster than any other model, while on the CIFAR100 ViT zoo 3-lvl-\mthd clearly outperforms all baselines. Unlike generalisation prediction, however, StatNN is highly competitive on ResNet18 when training data is scarce, achieving the best $\tau$ at 10\% of the training set, suggesting that layer-level statistics capture much of the signal relevant to robustness. \mthd overtakes it from $50\%$ of the training set onwards, where it achieves the best performance. As in generalisation prediction, the \RC{} variants perform on par with their GMN counterparts, while still outperforming all baselines on CIFAR100 ViT.

\begin{table}[h!]
\centering
\caption{mCE prediction. We report Kendall's $\tau (\uparrow)$, $R^2 (\uparrow)$, and convergence epoch (Ep.).}
\label{tab:mce}
\resizebox{\textwidth}{!}{%
\begin{tabular}{@{}l*{4}{ccc}@{}}
\toprule
Method &
\multicolumn{3}{c}{ResNet18-0.1} &
\multicolumn{3}{c}{ResNet18-0.5} &
\multicolumn{3}{c}{ResNet18} &
\multicolumn{3}{c}{CIFAR100 ViT} \\
\cmidrule(lr){2-4} \cmidrule(lr){5-7} \cmidrule(lr){8-10} \cmidrule(lr){11-13}
 & $\tau$ & $R^2$ & Ep. & $\tau$ & $R^2$ & Ep. & $\tau$ & $R^2$ & Ep. & $\tau$ & $R^2$ & Ep. \\
\midrule
StatNN 
& $\mathbf{0.847}$ & $0.913$ & $77$ & $0.864$ & $0.927$ & $92$ & $0.864$ & $0.934$ & $82$ & $0.899$ & $0.959$ & $\mathbf{86}$ \\
GMN & $0.808$ & $0.898$ & $87$ & $0.858$ & $0.930$ & $90$ & $0.867$ & $0.934$ & $87$ & $0.873$ & $0.918$ & $94$ \\
\midrule
SANE
& $0.586$ & $0.487$ & -- & $0.614$ & $0.583$ & -- & $0.632$ & $0.617$ & --& -- & -- & -- \\
\midrule
\midrule
2-lvl-\mthdT (Ours) & $0.798$ & $0.867$ & $83$ & $0.861$ & $0.929$ & $81$ & $\mathbf{0.870}$ & $\mathbf{0.942}$ & $82$ &  $0.925$ & $0.982$ & $98$ \\
2-lvl-\mthdT-\RC (Ours) & $0.783$ & $0.858$ &  $97$ &  $0.835$ & $0.918$ & $80$ & $0.865$ & $0.936$ & $85$ & $\mathbf{0.938}$ & $0.973$ & $95$   \\
3-lvl-\mthdT (Ours) & $0.831$ & $0.903$ & $72$ & $0.860$ & $0.922$ & $\mathbf{60}$ & $0.868$ & $0.938$ & $\mathbf{52}$ & $0.932$ & $\mathbf{0.985}$ & $96$\\
3-lvl-\mthdT-\RC (Ours) & $0.839$ & $\mathbf{0.918}$ & $\mathbf{72}$ & $\mathbf{0.882}$ & $\mathbf{0.937}$ & $74$ & $0.868$ & $0.939$ & $56$ & $0.914$ & $0.969$ & $92$  \\
\bottomrule
\end{tabular}
}
\end{table}

\noindent\textbf{Backdoor model detection.}
To mitigate backdoor vulnerabilities, recent literature increasingly analyses the footprint these attacks leave on NN parameters \citep{fields2021trojan, peng2025circumventing}. We evaluate whether metanetworks can detect backdoored models. We train 200 ResNet18 models on CIFAR10 for each of three attacks, BadNets \citep{gu2017badnets}, Blended \citep{chen2017targeted}, and WaNet \citep{nguyen2021wanet}, and consider both a binary task (per attack) and a multi-class task. We train by subsampling the majority "clean" class and report AUROC, accuracy, F1, and recall, which penalises poisoned models misclassified as clean.  2-lvl-\mthd (GRIT) instantiates $\mathfrak{F}_2$ as a GRIT \citep{ma2023graph} rather than a GMN as in 2-lvl-\mthd. Full details in \Cref{app:backdoor}.
As shown in \Cref{tab:backdoor}, \textbf{2-lvl-\mthd (GRIT) is the best-performing method across all settings}, almost perfectly distinguishing poisoned from clean models in every case (confusion matrices in \Cref{fig:backdoor-confusion}). StatNN 
performs strongly in some settings, even outperforming GMN, presumably because its simple statistical features are less prone to overfitting when training data is scarce. The advantage of \mthd over GMN suggests that detecting the traces left by backdoor attacks requires inductive biases that GMN does not easily capture, such as hierarchical or longer-range interactions. A more expressive layer alone does not explain this gap: on the Parameter Graph GMN (GRIT) performs on par with GMN, succeeding on BadNets but failing on Blended and WaNet.

\begin{table}[h!]
\centering
\caption{Backdoor model detection (averaged over 3 seeds - except GMN (GRIT)) on ResNet18.}
\label{tab:backdoor}
\resizebox{\textwidth}{!}{%
\setlength{\tabcolsep}{3.5pt}
\begin{tabular}{@{}l cccc cccc cccc cccc@{}}
\toprule
 & \multicolumn{4}{c}{BadNets} & \multicolumn{4}{c}{Blended} & \multicolumn{4}{c}{WaNet} & \multicolumn{4}{c}{Multiclass} \\
\cmidrule(lr){2-5} \cmidrule(lr){6-9} \cmidrule(lr){10-13} \cmidrule(lr){14-17}
Method & AUROC & Acc & F1 & Recall & AUROC & Acc & F1 & Recall & AUROC & Acc & F1 & Recall & AUROC & Acc & F1 & Recall \\
\midrule
StatNN
 & $0.986$ & $0.971$ & $0.830$ & $0.775$
 & $0.999$ & $0.991$ & $0.951$ & $0.978$
 & $0.997$ & $0.981$ & $0.896$ & $0.894$
 & $0.994$ & $0.959$ & $0.923$ & $0.899$ \\
GMN
 & $0.977$ & $0.954$ & $0.767$ & $0.832$
 & $0.454$ & $0.092$ & $0.170$ & $1.000$
 & $0.553$ & $0.503$ & $0.180$ & $0.566$
 & $0.959$ & $0.891$ & $0.791$ & $0.783$ \\

GMN (GRIT)
 & $0.978$ & $0.947$ & $0.748$ & $0.845$ 
 & $0.430$ & $0.818$ & $0.000$ & $0.000$
 & $0.578$ & $0.365$ & $0.180$ & $0.748$ 
 & $0.968$ & $0.917$ & $0.827$ & $0.794$ \\
 \midrule
 \midrule
2-lvl-\mthd (Ours)
 & $0.896$ & $0.913$ & $0.598$ & $0.664$
 & $\mathbf{1.000}$ & $0.995$ & $0.974$ & $0.970$
 & $0.970$ & $0.959$ & $0.775$ & $0.753$
 & $0.999$ & $0.983$ & $0.950$ & $0.948$ \\
2-lvl-\mthd (GRIT) (Ours)
 & $\mathbf{1.000}$ & $\mathbf{0.999}$ & $\mathbf{0.996}$ & $\mathbf{0.992}$
 & $0.999$ & $\mathbf{0.999}$ & $\mathbf{0.981}$ & $\mathbf{1.000}$
 & $\mathbf{1.000}$ & $\mathbf{0.999}$ & $\mathbf{0.992}$ & $\mathbf{0.995}$
 & $\mathbf{0.999}$ & $\mathbf{0.984}$ & $\mathbf{0.956}$ & $\mathbf{0.955}$ \\
\bottomrule
\end{tabular}%
}
\end{table}

\noindent\textbf{Pruning.} Pruning is a leading approach to compressing pre-trained networks \citep{yu2018nisp, wang2019eigendamage}. We consider unstructured pruning of individual weights \citep{dong2017learning}. Since mask selection is non-differentiable, we fix the mask with a magnitude ($\ell_1$) heuristic and train \mthd on the ResNet18 dataset to edit the unpruned parameters of $\fun$ to compensate for the pruned ones, minimizing a statistical distance between the outputs of the original and the pruned, edited network on held-out images. Full details, including the training objective, are provided in \Cref{app:pruning}.
As shown in \Cref{fig:pruning}, \mthd converges significantly faster than GMN while leading to smaller performance drops, particularly at intermediate pruning ratios, where the metanetwork can meaningfully compensate for the pruned weights. The exceptions are the extreme ratios, where there is little room for the metanetwork to intervene, and the two methods perform nearly identically.
Notably, these gains in both performance and convergence stem solely from a lightweight module operating on the coarsened graph. Since the coarsened graph is orders of magnitude smaller than the Parameter Graph, these layers add negligible computational overhead (see \cref{app:pruning}).

\begin{figure}[h!]
    \centering
    \includegraphics[width=1.0\textwidth]{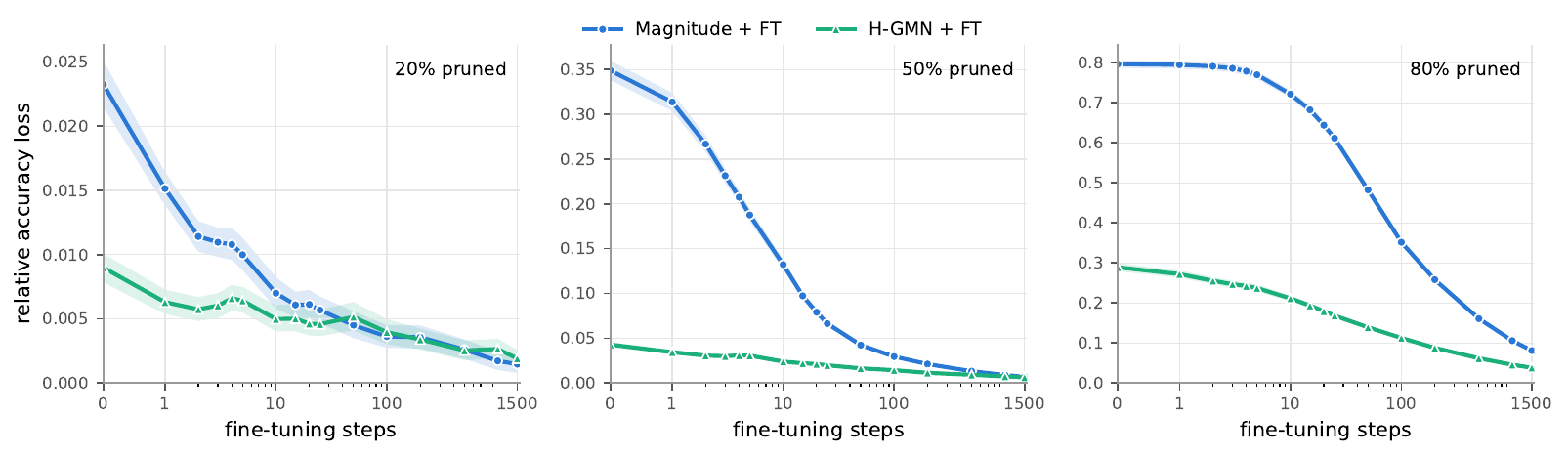}
    \caption{Fine-tuning after pruning: relative accuracy loss of the 1,200 test networks when fine-tuning from the magnitude-pruned weights (Magnitude + FT) or from the \mthd edit (\mthd + FT), at 80\%, 50\% and 20\% remaining weights. Step 0 is the one-shot result of \cref{fig:pruning}.}
    \label{fig:ft-pruned}
\end{figure}

\noindent\textbf{Comparison with fine-tuning.} A standard way to recover a pruned network is to fine-tune it. We fine-tune each pruned test network for up to 1500 steps, with the same mask and objective as the metanetwork, starting either from the magnitude-pruned weights or from the \mthd edit (details in \Cref{app:pruning}). As shown in \Cref{fig:ft-pruned}, fine-tuning from magnitude pruning needs about 10, 50 and 200 steps to match the one-shot \mthd edit at 80\%, 50\% and 20\% remaining weights. Used as the starting point of fine-tuning, the edit still halves the accuracy loss after 1500 steps at 20\% remaining weights, while at milder pruning both starting points reach the same accuracy.

\section{Conclusion and Limitations}
In this work, we propose \mthd, a graph metanetwork framework that processes NNs at multiple resolution levels, mirroring their hierarchical structure. Our experiments on ResNet18 and ViT models show that hierarchy is a valuable inductive bias and that operating on coarsened graphs substantially improves scalability without performance compromises. Regarding limitations, H-GMNs scale efficiently well beyond the models typically studied in the GMN literature, but there is still room for improvement concerning billion-parameter models. Additionally, symmetries beyond permutations have not been tested, although theoretically applicable.
Overall, with our work we aspire to draw attention to scalable, structure-aware weight space learning and to broaden the applicability of graph metanetworks to larger models.

\section*{Acknowledgements}
I.K. was supported by project MIS 5154714 of the National Recovery and Resilience Plan Greece 2.0 funded by the European Union under the NextGenerationEU Program. G.B. was partially funded by the Hellenic Foundation for Research and Innovation under the project ``ARCHNETS'' (4th Call for H.F.R.I. Research Project to support Postdoctoral Researchers, project no. 28415). This work was partially supported by a computational resources grant from The Cyprus Institute (HPC system ``Cyclone'') and from the European High Performance Computing Joint Undertaking (EuroHPC JU) and the Barcelona Supercomputing Center (BSC-CNS) (HPC system ``MareNostrum5 ACC'', project ID: EHPC-DEV-2026D02-289).

\subsection*{AI use statement}

In this work, we used generative AI tools 
to create or modify scientific figures or images, help with implementations and create or edit software code. Additionally,
to brainstorm certain methodological or experimental elements
and edit specific phrasings to improve readability. 
We have reviewed all AI-assisted work. For example,
all LLM-assisted code, including the implementation of our methods, was reviewed line by line, tested with unit tests and validated against reference implementations.  We
have verified the correctness of scientific figures, and reviewed and, where
necessary, revised all LLM-edited text.
We have not used generative AI tools to generate synthetic data sets, help develop theoretical models or conceptual frameworks, formulate mathematical claims, provide critical ingredients for proving mathematical claims, assist in the writing of proofs, propose or refine hypotheses, 
assist with translation, clean and reformat datasets, support qualitative and thematic data analysis, or interpret results.
 We take responsibility for the final content of this work,
including text, claims or artefacts produced with the aid of generative AI.

\subsection*{Ethics statement}

This work uses no human subjects or private data; We trained all metanetworks on publicly available benchmarks or on datasets that we created and will be released to the public. In our experimental evaluation, we use metanetworks for automated model evaluation and testing; these results require human oversight and should not be taken as a guarantee for model behaviour. Regarding backdoor detection, our results support weight-based auditing of third-party open-source models. However, the following should be stated: First, the backdoored models in our study were created solely for evaluation using standard poisoning protocols. Second, a published detector may also guide adaptive attacks designed to evade it, and therefore should be used with care. Third, detection results should not be taken as a guarantee that a model is safe. We do not foresee other direct negative societal impacts beyond those general to weight space learning methods.

\subsection*{Reproducibility statement}

The code for H-GMN, the baselines, and all training and evaluation scripts will be open-sourced. The methodology and model instantiations are described in detail in \cref{sec:method,sec:symmetries}, while graph construction, architectural details and 
hyperparameters are given in ~\cref{app:graph-constr,app:exp-details}. The datasets that we created
are described in Appendix~\ref{app:model-zoo-construction}.

\bibliography{iclr2027_conference}
\bibliographystyle{iclr2027_conference}

\appendix
\section{Appendix}
\subsection{Time and Memory Scaling}\label{app:time-mem}

\begin{figure}[h!]
    \centering
    \includegraphics[width=1.0\textwidth]{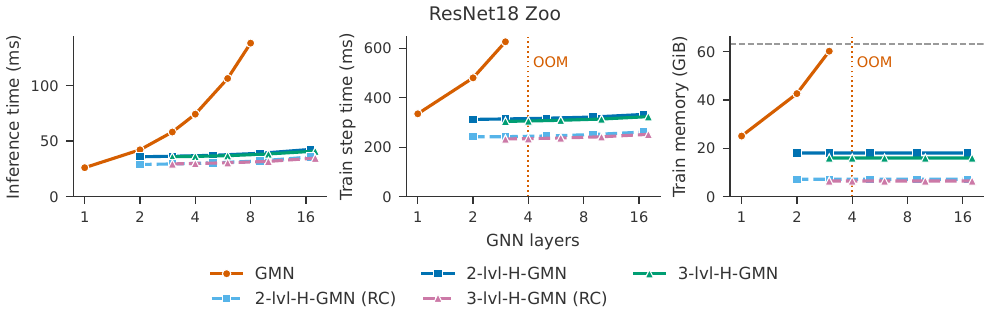}
    \caption{\textbf{ResNet18} Left: inference time per network ($B{=}1$). Middle and right: time and peak GPU memory of one training step ($B{=}4$). The x-axis counts all GNN layers. Dashed curves are the \RC{} variants. The dotted line marks where GMN runs out of memory, the dashed grey line is the GPU capacity (64\,GB).}
    \label{fig:resnet-mtn-depth}
\end{figure}

\begin{figure}[h!]
    \centering
    \includegraphics[width=1.0\textwidth]{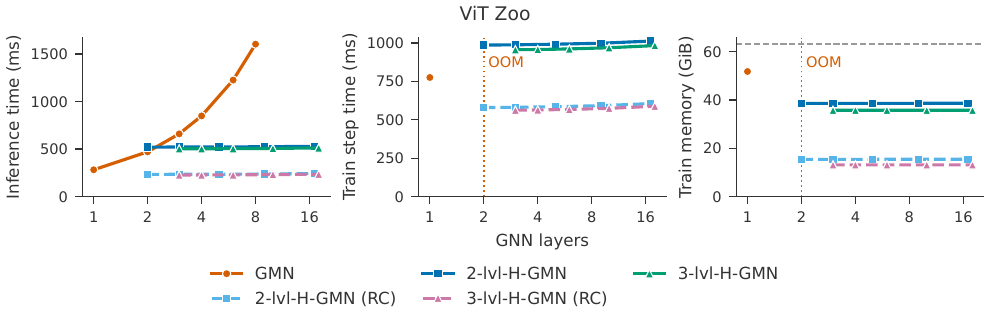}
    \caption{\textbf{CIFAR100 ViT} Left: inference time per network. Middle and right: time and peak GPU memory of one training step. $B{=}1$ throughout. Dashed curves are the \RC{} variants. The dotted line marks where GMN runs out of memory, the dashed grey line is the GPU capacity (64\,GB).}
    \label{fig:vit-mtn-depth}
\end{figure}

\begin{figure}[h!]
    \centering
    \includegraphics[width=1.0\textwidth]{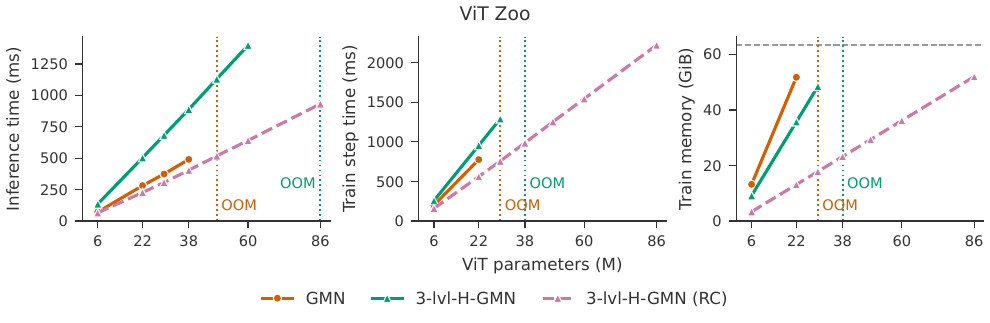}
    \caption{\textbf{Input network size (ViT)} Left: inference time per network. Middle and right: time and peak GPU memory of one training step, for randomly initialised ViTs of depth 12 and width 192--768. $B{=}1$ throughout; all metanetworks have $\sim$0.66M parameters and one layer per level. Dotted lines mark, in each model's colour, where it runs out of memory; the dashed grey line is the GPU capacity (64\,GB).}
    \label{fig:vit-input-size}
\end{figure}

We empirically verify the complexity analysis of \cref{sec:symmetries}. Using the notation of that section, for a flat GMN, both
memory  $O(L d^2 Kw)$ and time $O(L d^2 Kw^2)$ complexity scale linearly with the number of GMN's layers $K$. 
In contrast, for an H-GMN,
which has 
$O(Ld^2(k_1+1)w_1 + Lk_2w_2)$ memory 
and the $O(Ld^2(k_1+1)w_1^2 + Lk_2w_2^2)$ time complexity,
setting $k_1 = 1$, renders
the $d$-dependent term fixed, and additional layers only contribute to the second term of each complexity, which should be negligible.
\Cref{fig:resnet-mtn-depth,fig:vit-mtn-depth}
report inference time, training step time and peak training memory against the total number of GNN layers, for \mthdT{} with either a GMN or an \RC{} layer at the finest level (Appendix~\ref{app:rc}). \Cref{fig:vit-input-size} instead scales the input network at a fixed metanetwork depth.

\textbf{Setup.} All measurements are taken on a single NVIDIA H100 GPU (64\,GB), one configuration at a time, each in a separate process. For every configuration we discard 10 warm-up steps and report the median over the next 50 steps; timings are GPU-synchronised CUDA-event measurements of the metanetwork alone (data loading and host-to-device transfer are excluded), and memory is the peak allocated GPU memory during the timed steps. A training step comprises the forward pass, the loss, the backward pass, and an optimizer update. Inference time is reported per network ($B{=}1$); training steps use the per-GPU training batch size of each zoo ($B{=}4$ for ResNet18, $B{=}1$ for ViT). ResNet18 experiments run in fp32 and ViT experiments in bf16 mixed precision, matching the training setup of \cref{sec:exp}.
Hidden sizes are chosen so that all models have about the same number of parameters, as in the Lipschitz constant experiment; ResNet18: GMN $64$, 2-lvl-\mthd $58$, 3-lvl-\mthd $51$, each $\sim 0.9$M parameters; ViT: GMN $52$, 2-lvl-\mthd $38$, 3-lvl-\mthd $35$, each $\sim 0.66$M parameters. The \RC{} variants use the widths of their GMN counterparts, which changes their parameter count by less than $1\%$. We then vary depth at a single level while keeping all widths fixed: the number of layers of the flat GMN, $K \in \{1,2,3,4,6,8\}$, and the number of coarse-level layers of \mthd, $k_2 \in \{1,2,4,8,16\}$ (for 3-lvl-\mthd, the number of layers at the coarsest level), with a single Parameter Graph layer ($k_1{=}1$). Consequently, only the default-depth configurations are parameter-matched; parameters grow with depth. Once a configuration runs out of memory, deeper configurations of the same model are not run. We report memory only for training: during inference, the activations of each layer are released once the next layer has consumed them, so peak inference memory is nearly independent of depth (e.g.\ $3.5$\,GiB for the ResNet18 GMN from two to eight layers), whereas training must keep every layer's activations for the backward pass, which is what makes deep flat GMNs prohibitive.

\paragraph{ResNet18 zoo.} As shown in \cref{fig:resnet-mtn-depth}, GMN grows linearly with depth, as predicted, and runs out of memory beyond three layers. Both 2-lvl- and 3-lvl-\mthd variants (whose minimum total number of layers is 2 and 3, respectively) remain nearly constant up to 17 layers. This confirms that their cost is dominated by the single Parameter Graph layer. Replacing that layer with an \RC{} layer lowers the curves without changing their shape: training memory drops by $60\%$ (from $18.1$ to $7.2$\,GiB for 2-lvl-\mthd and from $16.0$ to $6.5$\,GiB for 3-lvl-\mthd), training step time by $23\%$ and inference time by $18$--$19\%$.

\paragraph{CIFAR100 ViT zoo.} Here $d$ is much larger, thus memory complexity $O(L d^2 Kw)$ makes a single GMN layer already expensive, and a second one exceeds GPU memory. As depicted in \cref{fig:vit-mtn-depth}, both \mthd variants again stay flat in all three metrics, since the coarse-level term is independent of $d$. The \RC{} layer matters more here than on ResNet18: it makes \mthd about $2.2\times$ faster at inference and $1.7\times$ faster per training step, with $60$--$63\%$ less training memory. At every depth we measured, both \RC{} variants are cheaper than a single-layer flat GMN (e.g.\ 3-lvl-\mthd: $227$ vs.\ $282$\,ms at inference; $563$ vs.\ $776$\,ms and $13.2$ vs.\ $51.8$\,GiB per training step).

\paragraph{Input network size.} We finally scale the input network instead of the metanetwork. We generate randomly initialised ViTs of depth 12 and width 192 to 768, i.e.\ from $6$M to $86$M parameters (ViT-B scale), and process them with the parameter-matched GMN, 3-lvl-\mthd and 3-lvl-\mthd (\RC{}) of the previous experiment, each with one layer per level, under the same protocol. Since the metanetworks are fixed, all costs grow linearly with the number of input parameters, as predicted by \cref{sec:symmetries}, and the models differ only in their slopes (\cref{fig:vit-input-size}). GMN runs out of memory beyond $38$M parameters at inference and $22$M in training, and 3-lvl-\mthd beyond $60$M and $29$M, respectively. 3-lvl-\mthd (\RC{}) is the only model that processes the $86$M-parameter ViT, both at inference and in training, on a single GPU.

\subsection{Graph Construction}\label{app:graph-constr}

The graph construction procedure is a map $\constr: \cF \to \cG$ that receives a network $\fun \in \cF$ and yields a graph $\constr(\fun) = G = (\cV, \cE, \xv, \xe)$. Within the Graph Metanetworks framework, $\constr$ is designed so that the resulting nodes and edges inherit the same symmetries as the network's own parameters. Although many of the modules we are mapping to the Parameter Graph have been addressed by \cite{lim2024graph}, our proposed method deviates in most cases. 
To construct the parameter $\cG$ of an input model $\fun$, we typically traverse the computational graph $C$ and progressively add new sets of nodes and edges with connectivity with the previous layers, dictated from each layer we encode. An abstract algorithm of this process is described in \Cref{alg:graph-construction}. In the following sub-sections we analyze how $\mathrm{ENCODE}$ parses.

\begin{algorithm}[h]
\caption{$\constr$: parameter-graph construction via layer-by-layer traversal of $C$}
\label{alg:graph-construction}
\begin{algorithmic}[1]
\Require NN $\fun$ with computational graph $C$, topologically ordered into layers $\ell_1, \dots, \ell_T$
\Ensure Parameter graph $G = (\cV, \cE, \xv, \xe)$
\State $G \gets (\emptyset, \emptyset)$
\State $\mathrm{active}[\text{input}] \gets$ one node per input channel of $C$
\For{$t = 1, \dots, T$} \Comment{visit $C$ in topological order}
    \State $U \gets \mathrm{active}[\cdot]$ for the tensor(s) feeding $\ell_t$
    \State $G, V_{\ell_t} \gets \Call{Encode}{\ell_t, U, G}$
    \State $\mathrm{active}[\mathrm{name}(\ell_t)] \gets V_{\ell_t}$
\EndFor
\State \Return $G$
\end{algorithmic}
\end{algorithm}

\noindent\textbf{Channel-mixing layers.} Let $\ell$ be a linear or convolutional layer with weight $\bW$ and bias $\mathbf b$ (taken to be $\mathbf 0$ when $\ell$ has none), and let $\cV$, with $|\cV| = d_{\text{in}}$, be its input vertices. $\mathrm{Encode}(\ell, \cV, G)$ adds one vertex per output channel and connects every input channel to every output channel:
\begin{align}
  \Delta \cV &= \cV_o, \; |\cV_o| = d_{\text{out}}, & \bx_V(j) &= [\mathbf b(j)], \\
  \Delta \cE &= \cV \times \cV_o, & \bx_E(i,j) &= \bW(j, i, :) \in \mathbb{R}^{k_\ell},
\end{align}
and returns $\cV_o$. The complete bipartite graph $\cV \times \cV_o$ is exactly the wiring rule already stated for channel-mixing layers in \cref{sec:pg-lim}. The only thing that varies across modules is the per-pair feature width $k_\ell$, the number of raw weight scalars $\ell$ allocates to a given (input, output) channel pair:
\begin{itemize}
  \item $k_\ell = 1$ for a \texttt{Linear} layer whose input width equals $d_{\text{in}} = |\cV|$, which is the ordinary scalar-weight case $\bx_E(i,j) = [\bW(j,i)]$;
  \item $k_\ell = \prod \text{kernel\_size}$ for \texttt{Conv1d/2d/3d}, where the edge carries the whole spatial kernel between the two channels (the sequence/spatial dimension does not enter the graph, as in the main text);
\end{itemize}

Attention's query/key/value/output projections are channel-mixing layers too, but a plain bipartite call is not enough to expose per-head structure to later pooling; they are treated separately in \cref{sec:graph-con,app:mhsa}. 

\noindent\textbf{Channel-wise layers.} Let $\ell$ be a normalization or pooling layer with input vertices $\cV$, $|\cV| = d_{\text{in}}$. Unlike channel-mixing layers, $\ell$ does not mix channels. It adds one new vertex per input channel, so $|\cV_o| = d_{\text{out}} = d_{\text{in}}$, and connects each input vertex \emph{one-to-one} to its output vertex instead of bipartitely. We index both sets by channel, so vertex $i \in \cV$ is connected to vertex $j = i \in \cV_o$:
\begin{align}
  \Delta \cV &= \cV_o, & \bx_V(j) &= [\beta(j)], \\
  \Delta \cE &= \{(i, j) \in \cV \times \cV_o : j = i\}, & \bx_E(i,j) &= [\gamma(j)],
\end{align}
and returns $\cV_o$. This is the channel-wise wiring rule of \cref{sec:pg-lim}. The values of $(\gamma, \beta)$ depend on the module:
\begin{itemize}
  \item \textbf{BatchNorm / InstanceNorm at inference} (running statistics tracked): the four per-channel quantities are folded into two effective affine terms,
  \begin{equation}
    \gamma(j) = \frac{\gamma^{\mathrm{raw}}(j)}{\sqrt{\sigma^2(j) + \epsilon}}, \qquad
    \beta(j) = \beta^{\mathrm{raw}}(j) - \gamma^{\mathrm{raw}}(j) \frac{\mu(j)}{\sqrt{\sigma^2(j) + \epsilon}},
  \end{equation}
  where $(\gamma^{\mathrm{raw}}, \beta^{\mathrm{raw}})$ are the module's own affine parameters and $(\mu, \sigma^2)$ its running mean and variance. This assumes a frozen checkpoint evaluated at inference, which is the trained-model setting that weight space learning already works in.
  \item \textbf{LayerNorm, GroupNorm, InstanceNorm without running statistics}: there are no running statistics to fold, so $\gamma(j) = \gamma^{\mathrm{raw}}(j)$ and $\beta(j) = \beta^{\mathrm{raw}}(j)$.
  \item \textbf{Pooling} (max/average, plain or adaptive): $\ell$ has no learnable parameters, so $\bx_V(j) = \bx_E(i,j) = [0]$. Which pooling operator was applied is recorded only in the edge type (\cref{app:pe}), not in $\bx_E$. Because standard pooling layers have no learnable parameters, their vertices and edges are identical for every model in a zoo of fixed architecture. In such setups they can be discarded without affecting representation quality.
\end{itemize}

\noindent\textbf{Skip connections.}\label{app:skip}
A skip connection adds two things to the graph. The first is an optional \emph{parametric} shortcut branch, e.g.\ a strided $1{\times}1$ convolution followed by BatchNorm, as in ResNet's downsample path. The second is the \emph{non-parametric} summation that recombines the shortcut with the main path. The shortcut branch needs no new rule: its channel-mixing and channel-wise layers are encoded exactly as in the previous two paragraphs. By the time $C$'s traversal reaches the sum, both operands therefore already have vertex sets in $G$. What remains is the sum itself.

Let $\ell$ be an elementwise addition that merges the main branch's vertices $\cV$ with the residual branch's vertices $\cV'$, where $|\cV| = |\cV'|$ so the two can be summed. We index both sets by channel. $\ell$ adds no new vertices. It connects the residual branch to the main branch with a single set of unweighted, one-to-one edges:
\begin{align}
  \Delta \cV &= \emptyset, \\
  \Delta \cE &= \{(i, j) \in \cV' \times \cV : j = i\}, & \bx_E(i,j) &= [1],
\end{align}
and returns $\cV$. Downstream layers continue from the main branch's vertices, and the residual branch's vertices are not touched again. When our model initializes the graph, vertices and edges that carry no parameters receive learnable embeddings instead of a fixed value. One embedding is shared by all vertices, and one by all edges, of the same layer.

\noindent\textbf{Multi-head self-attention.}\label{app:mhsa}
A multi-head self-attention block does not fit the plain channel-mixing rule. Restricted to permutations, its symmetry group is $\Sth^{\mathrm{MHSA}} = (S_{d_k} \times S_{d_v})^H \rtimes S_H$. This group permutes each head's query/key coordinates jointly and its value/output coordinates jointly, which a single bipartite edge set cannot express. Each head is therefore given its own pair of vertex sets.

Consider a block with $H$ heads,
$f_{C, \bth}(\bx) = \sum_{h=1}^{H} \bW^h_o (\bW^h_v \bx)\, \sigma\big((\bW^h_q \bx)^T (\bW^h_k \bx)\big)$,
with $\bW^h_q, \bW^h_k \in \mathbb{R}^{d_k \times d_{\text{in}}}$, $\bW^h_v \in \mathbb{R}^{d_v \times d_{\text{in}}}$, $\bW^h_o \in \mathbb{R}^{d_{\text{out}} \times d_v}$, and input vertices $\cV$, $|\cV| = d_{\text{in}}$ (in the notation of \cref{sec:graph-con}, $\cV = \mathbf V_l$). For each head $h \in [H]$, $\mathrm{Encode}$ adds two vertex sets, $\cV^h_k$ with $|\cV^h_k| = d_k$ and $\cV^h_v$ with $|\cV^h_v| = d_v$. It also adds one output set $\cV_o$ with $|\cV_o| = d_{\text{out}}$, shared by all heads. These sets are connected by four edge sets per head:
\begin{align}
  \cE^h_k    &= \cV \times \cV^h_k,     & \bx_E(i,j) &= [\bW^h_q(j,i), \bW^h_k(j,i)] \in \mathbb{R}^2, \\
  \cE^h_v    &= \cV \times \cV^h_v,     & \bx_E(i,j) &= [\bW^h_v(j,i)], \\
  \cE^h_{kv} &= \cV^h_k \times \cV^h_v, & \bx_E(i,j) &= [1/\sqrt{d_k}], \\
  \cE^h_o    &= \cV^h_v \times \cV_o,   & \bx_E(i,j) &= [\bW^h_o(j,i)],
\end{align}
so that $\cE_o = \bigcup_{h} \cE^h_o$ as in the main text, and
\begin{equation}
  \Delta \cV = \cV_o \cup \bigcup_{h=1}^H \big(\cV^h_k \cup \cV^h_v\big), \qquad
  \Delta \cE = \bigcup_{h=1}^H \big(\cE^h_k \cup \cE^h_v \cup \cE^h_{kv} \cup \cE^h_o\big).
\end{equation}
$\mathrm{Encode}$ returns $\cV_o$ ($= \mathbf V_{l+1}$) as the block's output vertices. The per-head sets $\cV^h_k, \cV^h_v$ remain in $G$ but are not passed on as input to later layers. As in the main text, biases are omitted for simplicity. $\cE^h_k$ and $\cE^h_{kv}$ get their own edge types, which distinguish them from linear layers; $\cE^h_v$ and $\cE^h_o$ use the linear-layer edge type.

$\cE^h_k$ stores the query and key weights of a channel pair together as one 2-dimensional feature. Any permutation of $\cV^h_k$ therefore reorders queries and keys together, which is exactly the action of $S_{d_k}$. $\cE^h_{kv}$ is complete bipartite with a constant, parameter-free feature. It links $\cV^h_k$ to $\cV^h_v$, so relabeling the heads ($S_H$) moves both sets of a head together.

\noindent\textbf{Positional encodings.}\label{app:pe}
A standard MPNN is equivariant to the full symmetric group $S_{|\cV|}$ acting on vertex indices: with no notion of vertex identity beyond adjacency and features, it treats \emph{any} relabeling of $G$ as a symmetry. $S_{|\cV|}$ is far larger than the group we actually want, $\Sth = \prod_g S_{|g|}$: it contains permutations that mix nodes across groups, which are not symmetries of $\fun$ and must not be treated as such. We therefore equip $G$, at construction time, with symmetry-breaking positional encodings $\bp_V, \bp_E$ that are shared within a group but differ across groups, restricting the graph's effective symmetry to exactly $\Sth$. This mirrors standard practice in graph metanetworks \citep{lim2024graph, kofinas2024graph, kalogeropoulos2024scale}.

Further, $\bp_V$ typically also carries each vertex's type and, where applicable, its activation function; similarly, $\bp_E$ carries each edge's type. Both remain constant within a group, so $\Sth$ is preserved.

\subsection{Model Zoo construction}\label{app:model-zoo-construction}

Weight space learning (WSL) is driven by the large number of trained models that are now publicly available: model hubs hold millions of checkpoints, and methods that treat these parameters as data could be used to audit, compare and edit models at scale. Benchmarking WSL methods has not kept pace. Most existing evaluations use populations of small MLPs or CNNs \citep{unterthiner2020predicting,schurholt2021self}, whose size and simplicity are far from the models people actually deploy. A few recent efforts have released populations of modern architectures, but each has a limitation that makes it a poor benchmark. The ResNet18 zoo of \cite{schurholt2022model} contains a checkpoint per training epoch, yet datapoint models differ only in the random seed. The training hyperparameters are held fixed, so the population covers a narrow range of model behaviour, and predicting a property from the weights becomes a question of seed noise rather than of real functional differences. A population of Vision Transformers has been proposed by \citep{falk2025model}, but only a
small fraction of their models has been released, too few to serve as a benchmark. To address both problems, we build four new model zoos: a ResNet18 zoo and a ViT zoo, both trained under deliberately varied hyperparameters, a zoo of ResNet18 backdoored models and a Varying ViT zoo with up to 86M parameters. Each checkpoint comes with a set of labels: clean accuracy, mean corruption error (mCE), and, for ResNet18, a Lipschitz constant upper bound. All datasets will be made publicly available. We describe each zoo below.

\subsubsection{ResNet18 Zoo}\label{app:model-zoo-construction-resnet}

\paragraph{Sampling strategy.}
Because every member of the zoo shares the ResNet18 architecture, structural
hyperparameters (depth, width, activation, normalisation) cannot be varied without
leaving the model family. All diversity in the population therefore has to come from
the training procedure. Every model uses the standard CIFAR variant of ResNet18 (${\sim}11.2$M parameters), is initialised with Kaiming-uniform initialisation ~\citep{he2015delving}, and is trained on CIFAR-10 for $50$ epochs with SGD (momentum $0.9$), batch size $128$, a one-cycle learning-rate schedule and standard augmentation (random crop with $4$-pixel padding and horizontal flips). For each model we independently sample the peak learning rate $\eta \sim \log\mathcal{U}[10^{-3}, 10^{-1}]$, the weight decay $\lambda \sim \log\mathcal{U}[10^{-6}, 10^{-2}]$, and the random seed, which controls initialisation and data ordering. Sampling on a logarithmic scale spreads the models evenly across two orders of magnitude in $\eta$ and four in $\lambda$. This range runs from strongly regularised to nearly unregularised training, and so produces populations that differ in accuracy, robustness and weight norm. 

\paragraph{Dataset size.}
The zoo contains $1{,}600$ ResNet18 models. We save a checkpoint every $5$ epochs,
giving $16{,}000$ checkpoints in total. Our experiments use the checkpoints at epochs
$\{10, 20, 30, 40, 50\}$, i.e.\ $8{,}000$ labelled samples. Models are split into
train, validation and test sets at the level of the training run ($1{,}120 / 240 / 240$
models), so all checkpoints of a model fall in the same split and no training
trajectory leaks across splits.

\subsubsection{ViT Zoo}\label{app:model-zoo-construction-vit}

Vision Transformers now dominate public model repositories, and in practice they are rarely trained from scratch: most checkpoints come from fine-tuning a small number of widely shared pretrained backbones. A realistic WSL benchmark for transformers should therefore capture both sources of variation, namely \emph{which} pretrained model a checkpoint starts from and \emph{how} it is fine-tuned. \citet{falk2025model} built a ViT zoo in this spirit, but only a small subset of its models has been released, which is too few to train and evaluate WSL methods. We therefore construct a new, larger ViT zoo, closely following their procedure.

\paragraph{Sampling strategy.}
As with ResNet18, the architecture is fixed: all models are ViT-S/16 (${\sim}21.7$M parameters), so diversity must come from the training pipeline. Following \cite{falk2025model}, we vary it along two axes. \emph{(i) Pretrained backbone.} Besides the two ViT-S/16 backbones of~\cite{falk2025model}, pretrained from scratch with a supervised and a contrastive objective respectively, we add three public \texttt{timm} checkpoints with different pretraining data (AugReg on ImageNet-1k, ImageNet-21k, and ImageNet-21k followed by ImageNet-1k fine-tuning \citep{steiner2022how}). This gives five backbones in total. \emph{(ii) Fine-tuning hyperparameters.} Using the fine-tuning protocol of~\citet{falk2025model}, each backbone is fully fine-tuned on CIFAR-100 (upsampled to $224\times224$, with random-crop and horizontal-flip augmentation) for $50$ epochs with a freshly initialised linear classification head. The optimiser is AdamW \citep{loshchilov2018decoupled} or SGD (Nesterov momentum $0.9$), with a constant learning rate $\eta \in \{10^{-4}, 3\cdot10^{-4}, 10^{-3}, 3\cdot10^{-3}\}$, weight decay $\lambda \in \{0, 10^{-5}, 10^{-4}, 10^{-3}\}$ and batch size $B \in \{64, 128, 256\}$. For each backbone we sample fine-tuning runs as follows: the random seed deterministically selects a configuration $(\eta, \lambda, B)$ from the resulting $4 \times 4 \times 3 = 48$-point grid, and every seed is fine-tuned with both optimisers. Across seeds, every backbone covers the full grid, and each (backbone, configuration, optimiser) combination is trained with one or two seeds, which control the initialisation of the new head and the data
ordering.

\paragraph{Dataset size.}
The zoo contains $728$ models: $160$ for each \texttt{timm} backbone and $124$ for each
backbone of~\citet{falk2025model}. We keep checkpoints at epochs
$\{10, 20, 30, 40, 50\}$, giving $3{,}640$ checkpoints. As for ResNet18, models are
split into train, validation and test sets at the model level ($70/15/15\%$), so all
checkpoints of a fine-tuning run fall in the same split.

\subsubsection{Varying ViT Zoo}\label{app:model-zoo-construction-varying-vit}
\paragraph{Sampling strategy.}
Unlike the ViT zoo, the architecture now varies, while the pretraining sources stay the same at every size. Pretrained checkpoints exist only for a few ViT sizes, and each size comes from different pretraining recipes, which would confound size with pretraining. We therefore derive every model from ViT-B/16 backbones and vary it along three axes. \emph{(i) Architecture.} Each model is a \texttt{timm} ViT/16 with embedding dimension $d \in \{192, 256, \dots, 768\}$ (head dimension $64$, i.e.\ $d/64$ heads), depth $L \in \{6, \dots, 12\}$ and MLP ratio $r \in \{2, 3, 4\}$, a grid of $210$ architectures. A child architecture is obtained from a ViT-B/16 parent by structured weight selection in the spirit of \citet{xu2024initializing}: it keeps the first $L$ blocks and, in each block, the $d$ residual channels, $d/64$ attention heads and $r d$ MLP units of largest magnitude. The $(768, 12, 4)$ cell is the parent itself, unsliced. \emph{(ii) Pretrained backbone.} We use eight public \texttt{timm} ViT-B/16 checkpoints: four AugReg models \citep{steiner2022how} (ImageNet-1k, ImageNet-21k, and two ImageNet-21k models fine-tuned on ImageNet-1k), the original ImageNet-21k ViT \citep{dosovitskiy2020image}, a SAM-trained model \citep{chen2021vision}, DINO \citep{caron2021emerging} and DeiT \citep{touvron2021training}.                     \emph{(iii) Fine-tuning hyperparameters.} Each model is fine-tuned on CIFAR-100 with the protocol of the ViT zoo (\Cref{app:model-zoo-construction-vit}): $50$ epochs, a freshly initialised linear head and a constant learning rate. The learning-rate grid now depends on the optimiser: AdamW uses $\eta \in \{10^{-5}, 3\cdot10^{-5}, 10^{-4}, 3\cdot10^{-4}\}$ and SGD (Nesterov momentum $0.9$) uses $\eta \in \{3\cdot10^{-4}, 10^{-3}, 3\cdot10^{-3}, 10^{-2}\}$. Weight decay and batch size use the grids above, giving $2 \times 4 \times 4 \times 3 = 96$ configurations. The runs of each architecture draw distinct configurations from this grid, and each run has its own seed.

\begin{figure}[h!]
    \centering
    \includegraphics[width=1.0\textwidth]{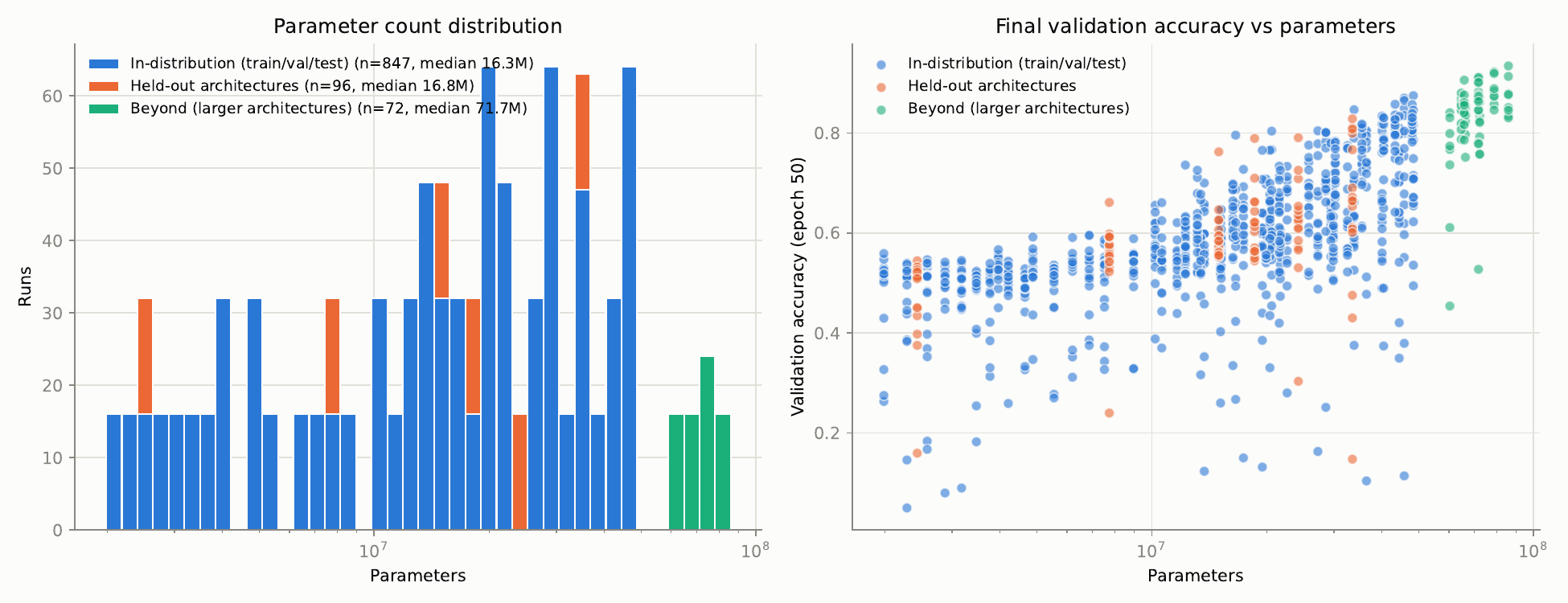}
    \caption{\textbf{Left}: Parameter count distribution of the Varying ViT zoo. \textbf{Right}: Validation accuracy at the last epoch $(50)$}
    \label{fig:varying-vit-params-ditrib}
\end{figure}

\paragraph{Dataset size.}
Before any training, we draw $59$ architectures with at most $48.4$M parameters, spread on a log scale down to the smallest grid cell ($2.0$M). We add all $9$ grid cells above $60$M parameters ($60.3$M--$85.9$M, up to ViT-B itself). No architecture between these two size bands is trained. Each of the $59$ smaller architectures is fine-tuned $16$ times (twice per backbone) and each larger one $8$ times (once per backbone), for $1{,}016$ runs. Runs diverged to non-finite weights got discarded, leaving $1{,}015$ models between $2.0$M and $85.9$M parameters. With checkpoints at epochs $\{10, 20, 30, 40, 50\}$, this gives $5{,}075$ checkpoints. Of the $59$ smaller architectures, $53$ are in-distribution. Their models are split at the model level into train, validation and test sets ($12/2/2$ runs per architecture; $636/105/106$ models). The remaining $6$ architectures ($96$ models, $2.4$M--$33.5$M parameters) form a test set of unseen architectures of familiar size, and the $9$ largest ($72$ models) form a test set of unseen, larger architectures. Backbones are balanced in every split, so only the architecture and its size change from one split to the next. In \cref{fig:varying-vit-params-ditrib}, we plot the distribution of model parameters, together with the validation accuracy at the last epoch $(50)$.

\subsubsection{Backdoor attacks}\label{app:backdoor}

\paragraph{Attacks.}
We consider three data-poisoning attacks with qualitatively different triggers. In all of them, a fraction $\rho$ of the training images is stamped with a trigger and relabelled to a target class $y_t$, and the model is trained as usual on the poisoned set. \emph{BadNets}~\citep{gu2017badnets} stamps an opaque white patch in the image corner. \emph{Blended}~\citep{chen2017targeted} blends the same patch into the image with weight $0.2$, which makes it less visible. \emph{WaNet}~\citep{nguyen2021wanet} replaces the additive patch with an imperceptible, smooth warping of the whole image, 
Our trigger implementations follow BackdoorBox~\citep{li2023backdoorbox}, a widely used open-source toolbox for backdoor learning, with its default CIFAR
configurations.

\paragraph{Construction.}
Every backdoored model is the \emph{twin} of a clean model in the corresponding zoo: it shares that model's architecture, training hyperparameters, random seed and checkpoint schedule, and differs only in the poisoned training data. For each model we sample the poisoning rate $\rho \sim \mathcal{U}[0.05, 0.5]$, a target class $y_t$ uniformly at random, and a trigger seed, which fixes the WaNet warping field. For ResNet18 we train $200$ backdoored twins per attack on CIFAR-10, using a $3\times3$ patch for BadNets and Blended. 
Each backdoored model inherits its twin's train/validation/test assignment, so a clean model and its backdoored counterpart never fall in different splits and a detector
cannot exploit shared hyperparameters across splits.

\subsection{Row--column (\RC{}) layer}
\label{app:rc}

\paragraph{Motivation.}
At the finest level, $\mathfrak{F}_1$ processes the full parameter graph and dominates the cost of an \mthdT{} (Appendix~\ref{app:time-mem}). A GMN layer applies an MLP to every edge, on its embedding concatenated with those of its endpoints, so each edge sees only its two endpoints. Standard message passing is not strictly required at this level, however: any layer that respects the permutation symmetries of the input network can instantiate $\mathfrak{F}_1$. The \RC{} layer, adapted from \citet{hartford2018deep}, updates each edge from its own embedding together with the mean embeddings of the edges sharing its output vertex (its row), its input vertex (its column) and its entire weight matrix, so that every weight sees its whole layer in a single step. The learned maps act on these means, of which there are about as many as vertices, rather than on every edge.

\paragraph{Notation.}
Let $G=(\cV,\cE,\hv,\he)$ be the parameter graph, with $\hv(v),\he(s,t)\in\R^{w}$, and $(\VL,\EL)$ its layer graph. Let $\gamma:\cV\to\VL$ map every vertex to its layer-graph vertex, i.e.\ to the layer it belongs to. For Transformers, all heads of a projection share one layer-graph vertex, so heads are not separated inside the \RC{} layer. Every layer-graph edge $\ell=(u,u')\in\EL$ corresponds to one weight matrix, whose entries are the edges
\begin{equation*}
    \cE_\ell=\{(s,t)\in\cE:\ \gamma(s)=u,\ \gamma(t)=u'\}.
\end{equation*}
Channel-wise layers and skip connections yield diagonal matrices. For example, in an MLP, $\cE_\ell$ holds the entries of the weight matrix $\bW_\ell$ of layer $\ell$, with $\xe(s,t)=\bW_\ell(t,s)$. The row of $t$ and the column of $s$ in $\cE_\ell$ are $\cN^{\mathrm{in}}_\ell(t)=\{s':(s',t)\in\cE_\ell\}$ and $\cN^{\mathrm{out}}_\ell(s)=\{t':(s,t')\in\cE_\ell\}$. For a vertex $v$, $\mathrm{in}(v)=\{\ell:\cN^{\mathrm{in}}_\ell(v)\neq\emptyset\}$ and $\mathrm{out}(v)=\{\ell:\cN^{\mathrm{out}}_\ell(v)\neq\emptyset\}$ are the layers whose matrices enter and leave $v$.

\paragraph{Statistics.}
For $\ell\in\EL$ and $u\in\VL$, the layer uses the row, column, matrix and vertex-group means
\begin{align}
    \b{r}_\ell(t) &= \frac{1}{|\cN^{\mathrm{in}}_\ell(t)|}\sum_{s\in\cN^{\mathrm{in}}_\ell(t)}\he(s,t), &
    \bc_\ell(s) &= \frac{1}{|\cN^{\mathrm{out}}_\ell(s)|}\sum_{t\in\cN^{\mathrm{out}}_\ell(s)}\he(s,t), \nonumber\\
    \bar{\bh}_E(\ell) &= \frac{1}{|\cE_\ell|}\sum_{(s,t)\in\cE_\ell}\he(s,t), &
    \bar{\bh}_V(u) &= \frac{1}{|\gamma^{-1}(u)|}\sum_{v\in\gamma^{-1}(u)}\hv(v).
    \label{eq:rc-stats}
\end{align}
The matrix and group means are mean pooling to the layer graph (eq.~\eqref{eq:pooling} with mean aggregation), and the updates below broadcast them back to the parameter graph, as in unpooling (eq.~\eqref{eq:unpooling}).

\paragraph{Updates.}
Each edge $(s,t)\in\cE_\ell$ is updated as
\begin{align}
    \he'(s,t) = \sigma\Big(&\bA_1\he(s,t) + \bA_2\bc_\ell(s) + \bA_3\b{r}_\ell(t) + \bA_4\bar{\bh}_E(\ell) \nonumber\\
    &{}+ \bA_5\hv(t) + \bA_6\bar{\bh}_V(\gamma(t)) + \bA_7\hv(s) + \bA_8\bar{\bh}_V(\gamma(s))\Big), \label{eq:rc-edge}
\end{align}
and each vertex $v$ as
\begin{align}
    \hv'(v) = \sigma\Big(&\bB_1\hv(v) + \bB_2\bar{\bh}_V(\gamma(v)) \nonumber\\
    &{}+ \sum_{\ell\in\mathrm{in}(v)}\big[\bB_3\b{r}_\ell(v) + \bB_4\bar{\bh}_E(\ell)\big] \nonumber\\
    &{}+ \sum_{\ell\in\mathrm{out}(v)}\big[\bB_5\bc_\ell(v) + \bB_6\bar{\bh}_E(\ell)\big]\Big), \label{eq:rc-vertex}
\end{align}
where $\bA_k,\bB_k\in\R^{w'\times w}$ are learnable (biases omitted for simplicity), $\sigma$ is an elementwise nonlinearity, and both updates are computed from the same inputs $\hv,\he$. Within a matrix the layer takes means; across matrices it takes sums, so every matrix entering or leaving $v$ contributes equally, whatever its fan-in or fan-out. The input projection, positional encodings, residual connections between layers and pooling are the same as with GMN layers.

\paragraph{Relation to \citet{hartford2018deep}.}
Restricted to a single matrix and without the vertex terms ($\bA_5,\dots,\bA_8$), eq.~\eqref{eq:rc-edge} is their exchangeable-matrix layer (Eq.~4), with means over the observed entries (their Sec.~2.1). The vertex terms couple the matrices that share vertices: vertex $s$ reads column $s$ of its outgoing matrices and row $s$ of its incoming ones (eq.~\eqref{eq:rc-vertex}), and feeds both back (eq.~\eqref{eq:rc-edge}). The layer is therefore not equivariant to permuting the rows and columns of each matrix independently, only to permutations consistent across matrices, as the symmetries of the network require. For a diagonal matrix, every row and column holds a single edge, so $\b{r}_\ell(t)=\bc_\ell(s)=\he(s,t)$, and the first line of eq.~\eqref{eq:rc-edge} reduces to $(\bA_1+\bA_2+\bA_3)\he(s,t)+\bA_4\bar{\bh}_E(\ell)$, a DeepSets layer over the channels.

\paragraph{Symmetry.}
Let $\pi$ be a permutation of $\cV$ that preserves the edge set and maps every vertex group onto itself, i.e.\ $(\pi(s),\pi(t))\in\cE\Leftrightarrow(s,t)\in\cE$ and $\gamma\circ\pi=\gamma$, acting on embeddings by $(\pi\cdot\hv)(v)=\hv(\pi^{-1}(v))$ and $(\pi\cdot\he)(s,t)=\he(\pi^{-1}(s),\pi^{-1}(t))$. Then $\pi$ maps every $\cE_\ell$ onto itself, rows onto rows and columns onto columns, while matrix and group means are unchanged. Every term of eqs.~\eqref{eq:rc-edge}--\eqref{eq:rc-vertex} is thus permuted with its edge or vertex, and the layer is equivariant to $\pi$. Every element of $\Sth$ permutes vertices within their layer-graph groups and preserves the graph, so the \RC{} layer respects the symmetries of the input network. As for GMN layers, positional encodings break the remaining permutations, which are not symmetries of $\fun$, e.g.\ those of the input or output neurons (Appendix~\ref{app:graph-constr}).

\paragraph{Cost.}
Consider the input MLP of \Cref{sec:symmetries}, with $L$ layers of width $d$, processed by a metanetwork of width $w$. Each weight matrix has $d^2$ edges but only $d$ rows and $d$ columns. Per edge, the \RC{} layer applies a single learned map, $\bA_1\he(s,t)$, at cost $O(w^2)$. All other learned maps act once per row, column, matrix or vertex, at cost $O(dw^2)$ per matrix, while computing the means and adding the resulting terms to the edges costs $O(d^2w)$. A layer therefore costs $O(Ld^2w^2)$ time and $O(Ld^2w)$ memory, as a GMN layer does, and the complexity analysis of \Cref{sec:symmetries} holds unchanged. The gain lies in the constants. Per edge, a GMN layer applies MLPs to concatenated inputs of width $2w$ (message) and $3w$ (edge update), whose first linear layers alone cost five $w\times w$ matrix--vector products, and it stores these concatenations for the backward pass. The \RC{} layer needs a single $w\times w$ product per edge and no concatenated inputs.

\subsection{Experimental details}\label{app:exp-details}
In this section, we provide additional experimental details regarding our experiments.

\noindent\textbf{Architecture.} Every \mthd applies a single message-passing layer on the Parameter Graph
($k_1 = 1$), since this is the only stage whose cost scales with the parameter count of the input
network. Vertex and edge pooling are implemented as a DeepSets aggregator \citep{zaheer2017deep}
over mean, max and standard deviation statistics \citep{corso2020principal}. On the coarsened graphs, the
networks $\mathfrak{F}_i$, $i \geq 2$, are instantiated either as message-passing GNNs or as GRIT
\citep{ma2023graph}. In the GRIT case, Relative Random Walk Probabilities (RRWP) are used to enlarge
the receptive field of each layer. The number of coarse-level layers is treated
as a hyperparameter. Positional encodings and the learnable embeddings of parameter-free vertices
and edges follow \cref{app:graph-constr}. The 2-lvl-\mthd variant pools the Parameter Graph
into the Layer Graph; for ResNet18 we additionally consider a 3-lvl-\mthd variant that further pools
into the ResBlock Graph, and for ViTs the Head Graph is used as an intermediate level.

\subsubsection{Lipschitz constant upper bound distribution.}\label{app:exp-details-lip}

\begin{figure}[h!]
    \centering
    \includegraphics[width=0.8\textwidth]{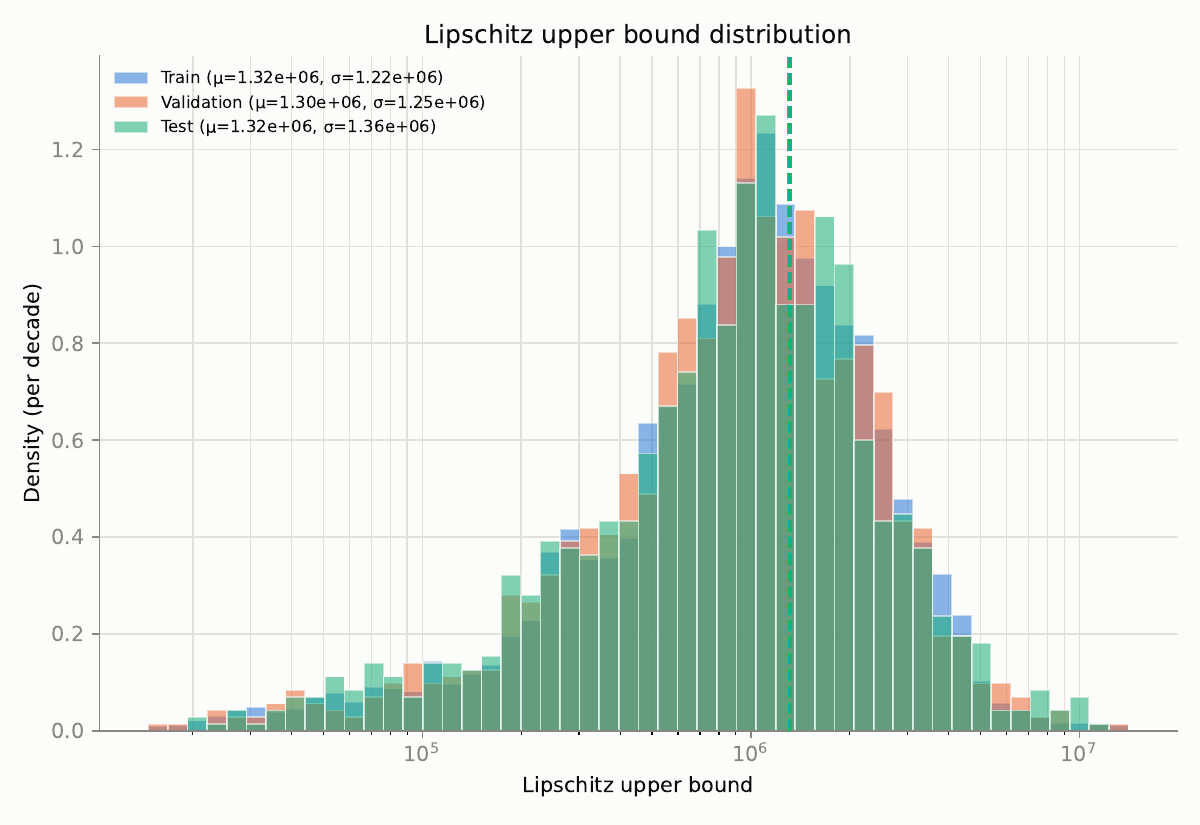}
    \caption{Distribution of the Lipschitz constant upper-bound proxy across the ResNet18 zoo.}
    \label{fig:resnet-lipschitz-ditrib}
\end{figure}

Unlike most experiments in the literature, where the ground-truth function is completely unknown, we propose a new experiment to evaluate the importance of hierarchy as an inductive bias. 
For ResNet-18 models, 
we bound the global network using the compositional product of its layers' spectral norms \citep{szegedy2013intriguing}. For residual blocks, following \cite{gouk2021regularisation}, we bound the block by the sum of the spectral norms of the main computational path and the skip connection. The $4$D convolution kernels are reshaped to $2$D matrices to compute their spectral norm, aligning with \cite{miyato2018spectral}. 
Since this reshaping does not guarantee to upper-bound the operator norm of a convolution, $f^\ast(\theta)$ should be read as a compositional proxy rather than a certified bound; what matters for our purposes is that it is an exact, known function of $\theta$ with an explicit hierarchical structure.
As a result, the ground-truth label for each input $\theta$ can be derived as:

\begin{equation} \label{eq:lipschitz}
    f^\ast(\theta) = \sigma_{\max}(\tilde{\bW}_{conv1}) \cdot \prod_{b=1}^{B} \left( s_b + \prod_{\ell=1}^{L_B} \sigma_{\max}(\tilde{\bW}_{b,\ell}) \right) \cdot \sigma_{\max}(\bW_{fc})
\end{equation}

where $B$ is the total number of residual blocks, $L_B$ the number of convolution layers within each block, and $s_b$ is the Lipschitz constant of the skip connection, defined as $s_b = 1$ for parameter-less identity shortcuts, and $s_b = \sigma_{\max}(\tilde{\bW}^{skip}_b)$ for downsampling projection shortcuts. The notation $\tilde{\bW}$ indicates the Conv-BatchNorm folded weights, defined as $\tilde{\bW}_{b,\ell} = \frac{\gamma_{b,\ell}}{\sqrt{\sigma_{b,\ell}^2 + \epsilon}} \bW_{b,\ell}$. We plot the distribution of those values in \cref{fig:resnet-lipschitz-ditrib}.

We plot the distribution of the Lipschitz constant upper bound in \Cref{fig:resnet-lipschitz-ditrib}. Evidently, our zoo consists of ResNet18 models of significantly varying Lipschitz constant upper bounds, with  $\sigma=1.22e+06$. \textbf{Setup}: In this experiment we aim to evaluate the hierarchy of neural networks as an inductive bias. Thus we perform our experiments on a very controlled setup. We fix the number of parameters and the number of GNN layers and change only the resolution on which each GNN layer is applied. Flat GMN applies 3 GNN layers on the Parameter Graph level. 2-lvl-\mthd uses 1 GNN layer in the fine-grained resolution, applies pooling of symmetric statistics \citep{corso2020principal} and then 2 GNN layers on the coarsened graph. Finally, 3-lvl-\mthd spans 3 different hierarchy levels with one GNN layer in each. The second pooling need not be a permutation invariant aggregation, since, in ResNet models, $\mathrm{Pool}_1$ has already collapsed all parameter symmetries. We simply concatenate the node/edge features of the elements that reside in the same residual block. 
Downsampling blocks contain two more layers than identity blocks (the $1\times1$ convolution and BatchNorm of the projection shortcut), so their concatenated features are longer. We therefore use a separate projection for each block type, mapping both to a common dimension. For datasets with heterogeneous architectures, where such mismatches are more frequent, zero-padding to a fixed number of slots, or an aggregation over the block's layers combined with positional encodings that identify each layer, would scale better.

\begin{figure}[h!]
    \centering
    \begin{subfigure}[h]{0.49\textwidth}
        \centering
        \includegraphics[width=\linewidth]{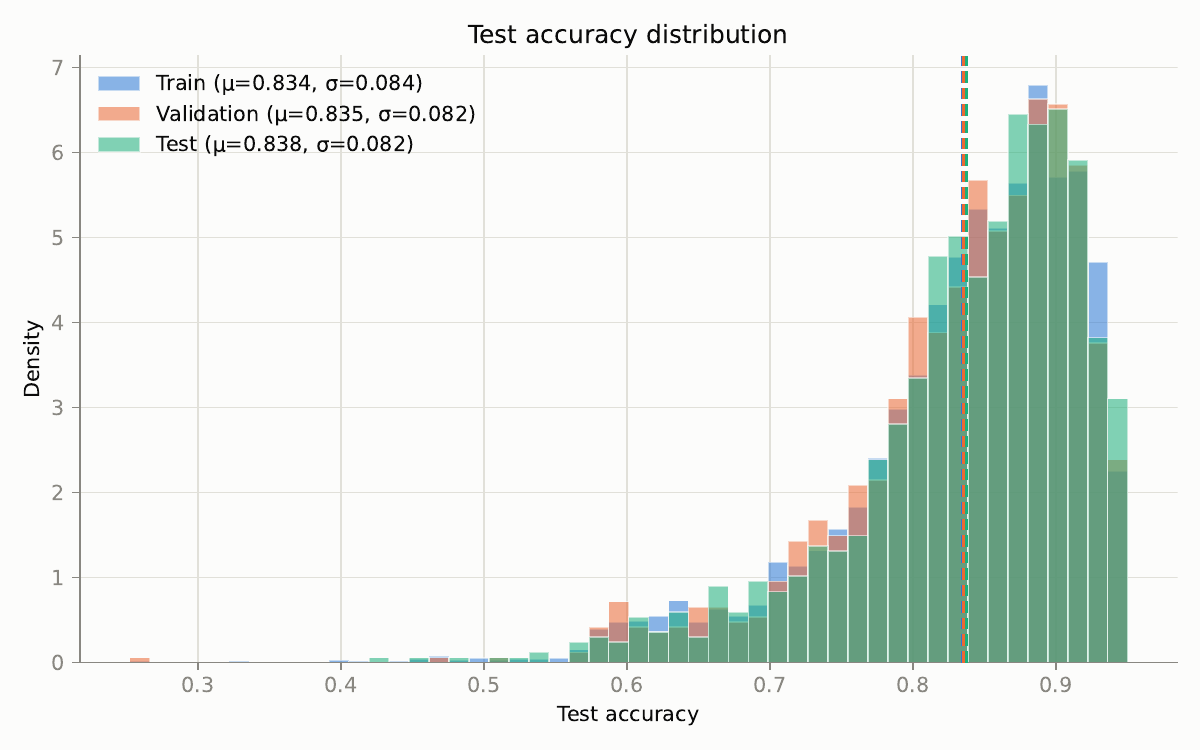}
        \caption{Test accuracy}
        \label{fig:resnet18-acc-distrib}
    \end{subfigure}\hfill
    \begin{subfigure}[h]{0.49\textwidth}
        \centering
        \includegraphics[width=\linewidth]{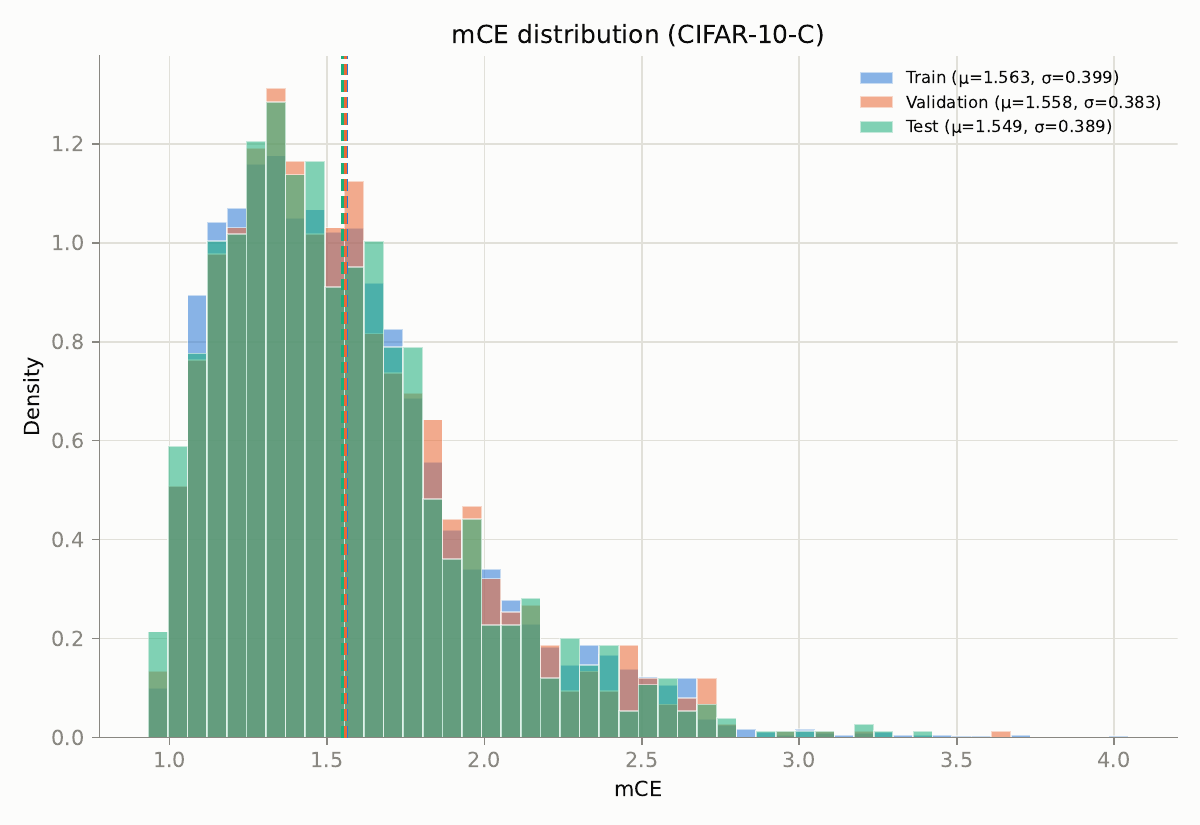}
        \caption{mCE}
        \label{fig:resnet18-mce-distrib}
    \end{subfigure}
    \caption{Distributions of (a) test accuracy and (b) mean corruption error (mCE) across the ResNet18 zoo.}
    \label{fig:resnet18-acc-mce-distribs}
\end{figure}

\begin{figure}[t]
    \centering
    \begin{subfigure}[t]{0.49\textwidth}
        \centering
        \includegraphics[width=\linewidth]{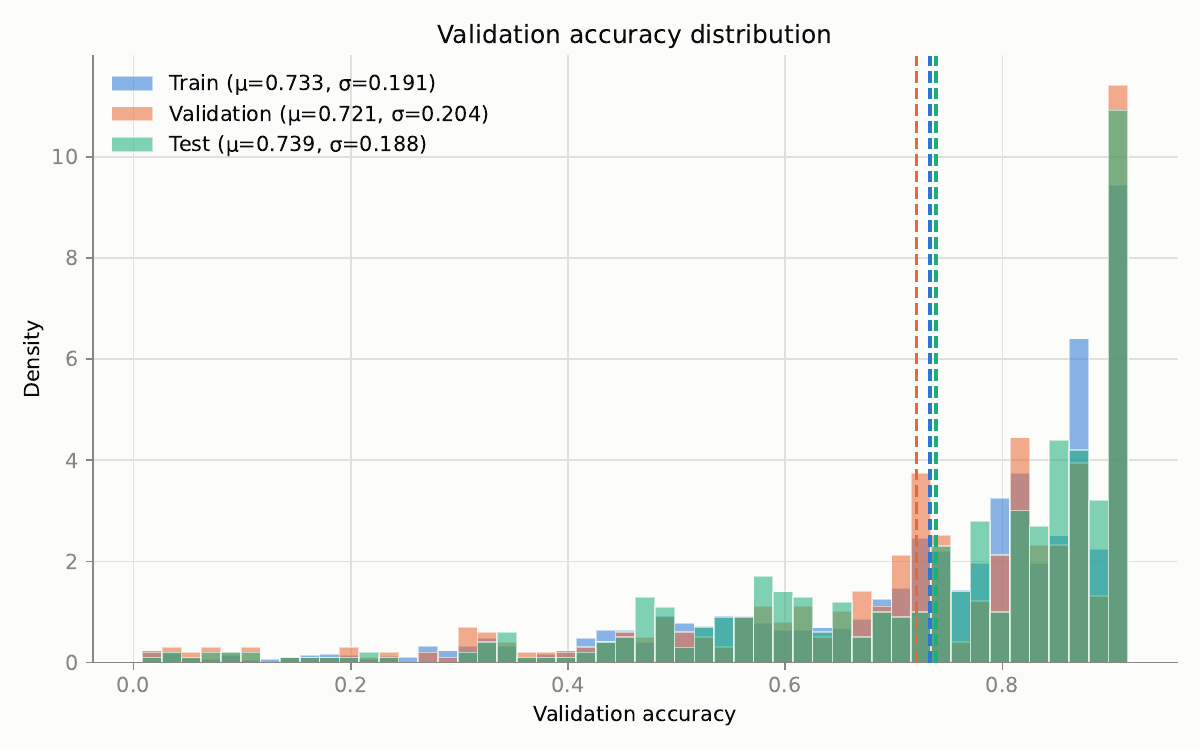}
        \caption{Validation accuracy}
        \label{fig:vit-val-acc-distrib}
    \end{subfigure}\hfill
    \begin{subfigure}[t]{0.49\textwidth}
        \centering
        \includegraphics[width=\linewidth]{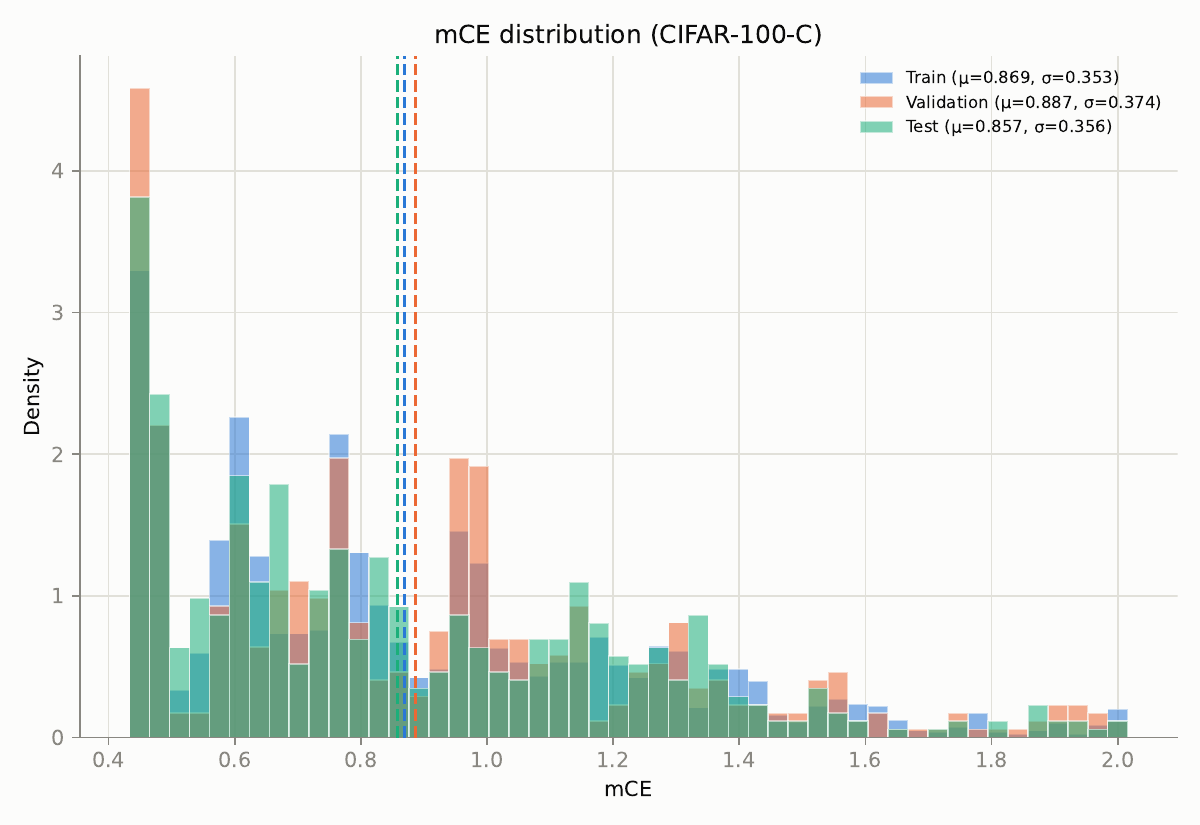}
        \caption{mCE}
        \label{fig:vit-mce-ditrib}
    \end{subfigure}
    \caption{Distributions of (a) validation accuracy and (b) mean corruption error (mCE) across the CIFAR100 ViT Zoo.}
    \label{fig:vit-distribs}
\end{figure}

\subsubsection{Generalisation prediction}\label{app:exp-details-gen-pred}
\paragraph{ResNet18 Zoo}
We plot the distribution of test accuracy in \Cref{fig:resnet18-acc-distrib}. Evidently, our zoo consists of ResNet18 models of significant variety. \textbf{Hyperparameters}: For \mthd we use only one GNN layer at the Parameter Graph resolution. In the coarsened graph, typically denoted as 2-lvl-\mthd, we experiment with both Graph Neural Networks and GRIT \citep{ma2023graph} for the instantiation of $\mathfrak{F}_2$. In both cases, we treat the number of layers as a hyperparameter and search the number of layers $k_2 \in \{ 1,2,3\}$. In the GRIT case, we also use Relative Random Walk Probabilities (RRWP) to swiftly increase the receptive field of each layer \citep{ma2023graph}. Additionally, we set batch size $=4$, hidden dimension for node/edge features in $\{32, 64\}$ for $\mathfrak{F}_1$ and in $\{64, 128\}$ for $\mathfrak{F}_2$. We also search learning rate in $\{0.001,0.0005,0.0001\}$ and weight decay in $\{0.0,0.001\}$. After converging in a specific configuration, we then run the same experiment for all training size options. For the GMN baseline, we search $k_1 \in \{2, 3, 4 \}$ and the hidden size in $\{ 32, 64, 128\}$ for the combinations that are able to fit in a GPU. The rest of the hyperparameters are treated the same way as in \mthd. For SANE \citep{schurholt2024towards}, to avoid costly re-pre-training for each training subset, we reuse the embeddings obtained from pre-training on the full training dataset and train only the downstream head on the corresponding subset. Despite this advantage, SANE underperforms all other methods.

\paragraph{CIFAR100 ViT Zoo}
In this Zoo, we use as target values for the generalisation prediction task, the accuracy on the validation split \citep{falk2025model}. The distribution across the ViT Zoo is shown in \Cref{fig:vit-val-acc-distrib}.
\textbf{Hyperparameters}: Again, H-GMN applies a single GNN layer on the Parameter Graph ($k_1=1$). On the
coarsened graphs, $\mathfrak{F}_2$ is a GMN on the Head Graph and $\mathfrak{F}_3$ a GRIT~\citep{ma2023graph} with
RRWP on the Layer Graph. We set the batch size to 1 and use distributed data parallelism based on
\texttt{torch.distributed}. We search the numbers of layers $k_2$ and $k_3$, the hidden dimensions of the node/edge
features of $\mathfrak{F}_1$ and $\mathfrak{F}_2$, and the learning rate in $\{0.001, 0.0005, 0.0001\}$, while fixing
the hidden dimension of $\mathfrak{F}_3$ to 256 and the weight decay to 0. We select $k_2=2$, $k_3=4$, hidden
dimensions of 52 and 128 for $\mathfrak{F}_1$ and $\mathfrak{F}_2$, and a learning rate of 0.0001. For the GMN
baseline, we use $k_1=1$, since a second layer does not fit in GPU memory, and a hidden dimension of 52. The rest of
the hyperparameters are treated the same way as in H-GMN.

\paragraph{MNIST ViT Zoo}
We follow the same procedure for the MNIST ViT zoo~\citep{tran2025equivariant}, with a few
differences. Since the input networks are much smaller, deeper flat GMNs fit in GPU memory: for the GMN baseline, we report the configuration of the depth ablation in \cref{app:abl-deeper-mnist-vit} that performs best on validation, i.e.\ $k_1=8$ with a hidden dimension of 128, trained with the same learning rate, number of epochs and effective batch size as \mthd (via gradient accumulation). We evaluate the Transformer-NFN baseline~\citep{tran2025equivariant} only on this dataset, since applying it to the $\sim$21M-parameter CIFAR100 ViT is intractable: its cost scales cubically with the embedding dimension of the input network and quadratically with
the channel width of the metanetwork.

\begin{figure}[h!]
    \centering
    \includegraphics[width=0.8\textwidth]{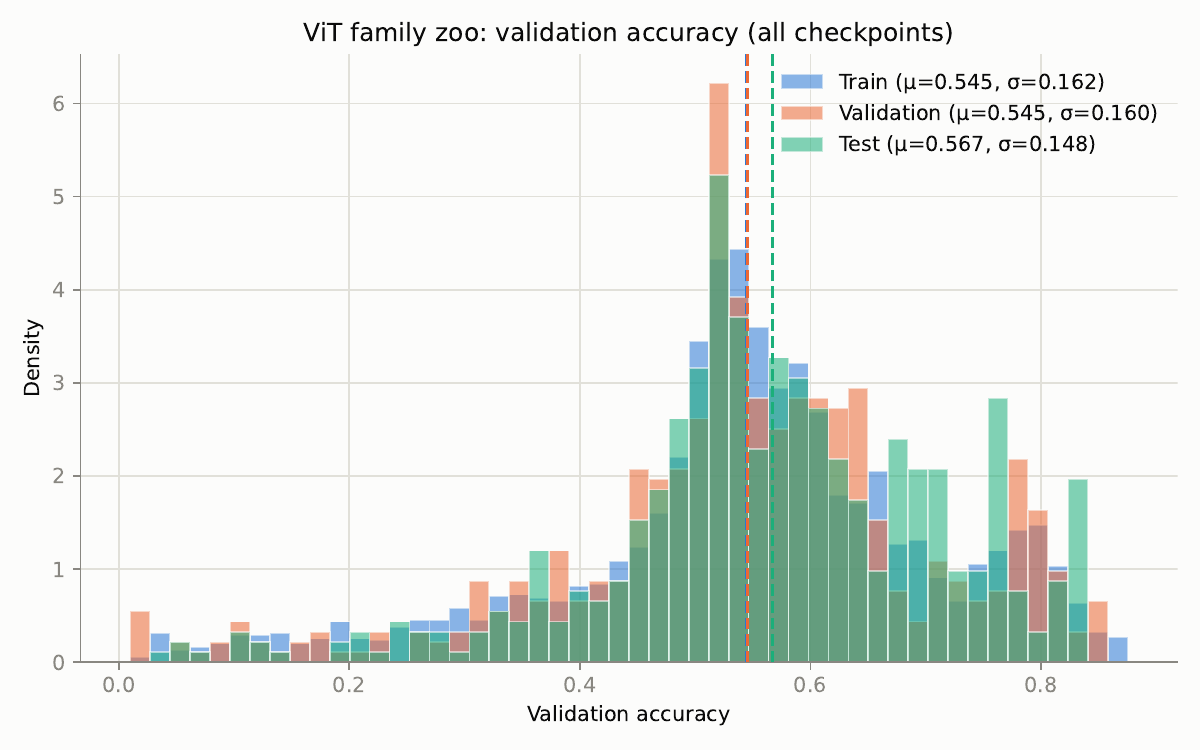}
    \caption{Distribution of the validation accuracy across the Varying ViT zoo.}
    \label{fig:varying-vit-acc-distrib}
\end{figure}

\paragraph{Varying ViT Zoo}
Similarly in this Zoo, we use as target values for the generalisation prediction task, the accuracy on the validation split. The distribution across the Varying ViT zoo is shown in \Cref{fig:varying-vit-acc-distrib}.
\textbf{Hyperparameters}: Again, for \mthd we use only one \textit{RC} layer at the Parameter Graph resolution. In the coarsened graph, we use 2 GMN layers for $\mathfrak{F}_2$ (width 128), then 4 GRIT layers with RRWP for $\mathfrak{F}_3$ (width 256). Additionally, we set batch size $=1$ and use distributed data parallelism based on $\mathrm{torch.distributed}$. We also use a learning rate equal to $0.0001$ and $0.0$ weight decay. For the GMN baseline, we search $k_1 \in \{1,2,3 \}$ with the hidden size decreased accordingly to fit into the GPU. The rest of the hyperparameters are treated the same way as in \mthd.

\subsubsection{Mean Corruption Error Prediction}\label{app:exp-details-mce}
Predicting a checkpoint's robustness to distribution shift is a natural task for graph metanetworks. Following \citet{hendrycks2019benchmarking}, we measure robustness by the mean Corruption Error (mCE), which normalizes the error on each corruption type by that of a fixed reference model before averaging, and use it as the regression target for the metanetwork. For each model zoo, we compute mCE over the $|C| = 19$ corruption types of CIFAR-10-C or CIFAR-100-C (the 15 standard corruptions plus four additional ones introduced to test generalisation beyond them) and five severity levels as shown in \cref{eq:mce-formula}:

\begin{equation}\label{eq:mce-formula}
CE_c^f = \frac{\sum_{s=1}^{5} E_{s,c}^f}{\sum_{s=1}^{5} E_{s,c}^{\mathrm{ref}}}, \qquad
\mathrm{mCE}^f = \frac{1}{|C|}\sum_{c=1}^{|C|} CE_c^f.
\end{equation}

The reference model is a single ResNet18 trained from scratch on CIFAR-10/CIFAR100. Since it is shared across all models in the zoo, the normalization only rescales each corruption type's contribution by a constant and does not affect the relative ordering of models on any individual corruption.

\paragraph{ResNet18 Zoo}
For the mean corruption error, we use the CIFAR-10-C dataset \citep{hendrycks2019benchmarking}. The distribution of mCE is shown in \Cref{fig:resnet18-mce-distrib}.
\textbf{Hyperparameters.} For both \mthd variants and GMN, we use the same hyperparameter search as for the ResNet18 zoo (\cref{app:exp-details-gen-pred}). For SANE \citep{schurholt2024towards}, we reuse the embeddings from pre-training on the full dataset and train the downstream head on the corresponding training subset.

\paragraph{CIFAR100 ViT Zoo}
For the CIFAR100 ViT zoo, we measure robustness as the mCE on CIFAR-100-C \citep{hendrycks2019benchmarking}, whose distribution is shown in \Cref{fig:vit-mce-ditrib}. As for ResNet18, the hyperparameter search for both \mthd and GMN follows \cref{app:exp-details-gen-pred}.

\subsection{Backdoor Detection}\label{app:exp-details-backdoor}

We use the backdoored ResNet18 zoo described in \Cref{app:backdoor}, in which every poisoned model shares its train/validation/test split with its clean counterpart. We consider four classification tasks: three binary tasks (clean vs.\ BadNets, clean vs.\ Blended, and clean vs.\ WaNet) and a multi-class task that distinguishes clean models from each of the three attacks.

Since clean models far outnumber poisoned ones, we apply random subsampling to obtain balanced classes in the training split. Validation and test splits are kept intact. We compare five methods: StatNN \citep{unterthiner2020predicting}, the flat GMN and GRIT baselines, and two variants of 2-lvl-H-GMN. The flat GRIT baseline applies $k_1=3$ GRIT layers on the Parameter Graph, without RRWP. Both variants apply a single GMN layer on the Parameter Graph ($k_1 = 1$) before pooling to the Layer Graph, on which $\mathfrak{F}_2$ is either a GMN or GRIT with RRWP \citep{ma2023graph} (2-lvl-\mthd-GRIT). As before, the hyperparameter search closely follows \cref{app:exp-details-gen-pred} for the ResNet18 zoo.

\begin{figure}[htbp]
    \centering
    \begin{subfigure}[b]{0.24\textwidth}
        \centering
        \includegraphics[width=\textwidth]{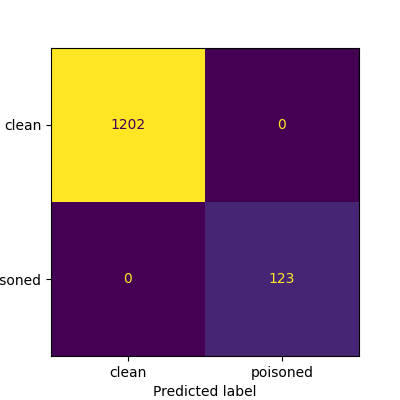}
        \caption{Badnets VS clean}
        \label{fig:img1}
    \end{subfigure}
    \hfill
    \begin{subfigure}[b]{0.24\textwidth}
        \centering
        \includegraphics[width=\textwidth]{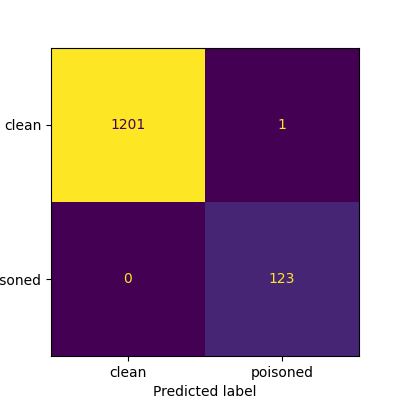}
        \caption{Blended VS clean}
        \label{fig:img2}
    \end{subfigure}
    \hfill
    \begin{subfigure}[b]{0.24\textwidth}
        \centering
        \includegraphics[width=\textwidth]{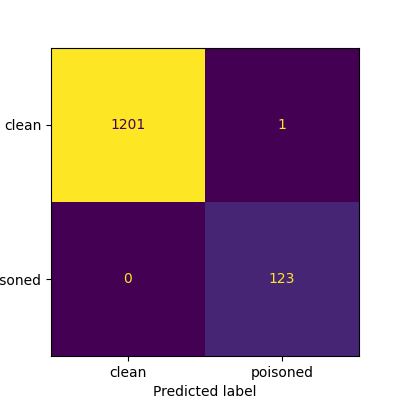}
        \caption{Wanet VS clean}
        \label{fig:img3}
    \end{subfigure}
    \hfill
    \begin{subfigure}[b]{0.24\textwidth}
        \centering
        \includegraphics[width=\textwidth]{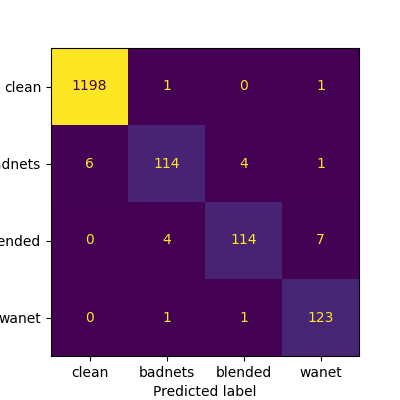}
        \caption{Multiclass}
        \label{fig:img4}
    \end{subfigure}
    
    \caption{Confusion matrices of 2-lvl-\mthd (GRIT) predictions on each backdoor detection task.}
    \label{fig:backdoor-confusion}
\end{figure}

\textbf{Results.} We select the checkpoint with the best validation F1 score. As shown in \Cref{tab:backdoor}, 2-lvl-\mthd (GRIT) is the strongest detector overall, almost perfectly separating poisoned from clean models for every attack, including the multi-class setting (see the confusion matrices in \Cref{fig:backdoor-confusion}). Both flat baselines detect BadNets but collapse on Blended and WaNet, whose triggers are less conspicuous. StatNN remains a strong baseline throughout, suggesting that simple layer statistics already carry much of the backdoor signal. 2-lvl-\mthd matches or outperforms StatNN on Blended and the multi-class task but falls behind on BadNets and WaNet. Instantiating $\mathfrak{F}_2$ with GRIT \citep{ma2023graph} closes this gap, hinting that long-range interactions between layers are important for this task.

\subsection{Pruning}\label{app:pruning}

Towards complying with compute and energy requirements, numerous techniques have been developed to compress pre-trained networks \citep{hinton2015distilling, han2015learning, wu2016quantized}, among which network pruning stands out as a leading approach \citep{yu2018nisp, wang2019eigendamage}. 
We consider unstructured pruning of individual weights \citep{dong2017learning} and train metanetworks to perform it via editing.
However, to circumvent non-differentiability issues of mask selection, we derive the pruning mask via a simple heuristic, namely the $\ell_1$ norm of the weights. The metanetwork's task is then to edit the unpruned parameters of the input network $\fun$ to compensate for the pruned ones. \mthd is thus instantiated as described in \cref{eq:decode}, with the following training objective:
\begin{equation}
\min_{\tilde{\bth}}
\bigg(
\underset{\sampling}{\mathbb{E}}
    \text{d}\Big(
            f_{C, \bm{\theta}}(\bx), 
            f_{C, \tilde{\bm{\theta}}}(\bx)
        \Big)
\bigg),
\end{equation}

where $\sampling$ denotes labeled images not seen by $\fun$ during training, and $\mathrm{d}$ is a statistical distance, $\tilde{\bm{\theta}} = \mathfrak{F}_{\mathrm{edit}}(\fun)$ as described in \Cref{sec:method}.
In our experiments, we used Jensen–Shannon divergence \citep{61115}.
At inference time, $\mathfrak{F}_{\mathrm{edit}}$ requires access only to the network parameters $\bth$. Since the pruning mask is computed by the same heuristic, \mthd and GMN face exactly the same task at each pruning ratio: editing the same set of unpruned parameters so that the edited model remains functionally close to the original.

For the pruning experiment, we utilize the editing pipeline described in \Cref{sec:method}. We apply 1 GNN layer at the Parameter Graph resolution and PNA-style \citep{corso2020principal} pooling to obtain the coarsened graph. The $\mathfrak{F}_2$ model is then instantiated as GRIT \citep{ma2023graph} coupled with Relative Random Walk Probabilities, to achieve maximum receptive field. Finally, we employ the UnPooling operation of \Cref{eq:decode} and one final GNN layer on the fine-grained resolution for the $\tilde{\mathfrak{F}}_1$ network. Both $\mathfrak{F}_1$ and $\tilde{\mathfrak{F}}_1$ are instantiated as single-direction GMNs in contrast to editing approaches followed by \cite{kalogeropoulos2024scale, kofinas2024graph}. 

To obtain the final weights in $\mathfrak{F}_{\mathrm{edit}}(\fun) = \constr^{-1}\big(\psi(\tilde\bh^{(1)})\big)$, we implement $\psi$ as a residual update: the metanetwork output is scaled by learnable per-layer scalar multipliers and added to the input $\bh^{(0)} = \mathfrak{G}(\fun)$. The metanetwork thus predicts a small correction to the input parameters instead of the new parameters themselves, following prior work \citep{zhou2023permutation, kofinas2024graph, kalogeropoulos2024scale}. Moreover, to keep only the remaining parameters, $\psi$ additionally applies the pruning mask to the updated parameters, so that pruned weights remain zero and only the unpruned ones are edited.

\textbf{Hyperparameters}: For \mthd we only use 2-lvl-\mthd and search for the number of layers in the coarsened level $k_2\in \{1,2,3\}$. For the GMN baseline, we search $k_1 \in \{2,3,4\}$. For both of them, we search learning rate in $\{0.001,0.0005,0.0001\}$ and set fixed batch size of 4. After converging in a specific configuration, we then run the same experiment for all pruning intensity options. 
In our experiments, the best GMN configuration used two GNN layers ($k_1 = 2$). Even in this setup, where \mthd and GMN have the same number of layers on the Parameter Graph, the additional cost of $\mathrm{Pool}_1$, $\mathfrak{F}_2$, and $\mathrm{UnPool}_1$ is negligible: inference takes $60$\,ms per network for \mthd versus $58$\,ms for GMN.

\subsubsection{Editing VS conventional recovery}

To compare editing with conventional recovery (\Cref{fig:ft-pruned}), we fine-tune every pruned test network (1,200 networks: 240 training runs, 5 checkpoints each) at 20\%, 50\% and 80\% pruning, starting either from the magnitude-pruned weights (Magnitude + FT) or from the \mthd edit (\mthd + FT). Both use the same magnitude mask, which stays fixed: pruned weights receive no gradient and are re-masked after every step. The loss is the metanetwork's objective, the Jensen--Shannon divergence between the outputs of the dense and the pruned network, so no labels are used. We train exactly the parameters the metanetwork edits. We use SGD with momentum $0.9$, no weight decay and a constant learning rate, on batches of 128 CIFAR10 training images with random crops and horizontal flips. Both starting points of a network see the same batches. We report the relative accuracy loss on the same test images as in \Cref{fig:pruning}, at 16 steps between 0 and 1500.

\paragraph{Learning rate.}
We select the learning rate from $\{10^{-4}, 3\cdot10^{-4}, 10^{-3}, 3\cdot10^{-3}, 10^{-2}\}$ on 16 validation networks spanning the zoo's training learning rates, fine-tuning from the magnitude-pruned weights only, at 20\% and 80\% pruning. We pick the rate with the lowest JSD on 256 held-out CIFAR10 training images (excluded from fine-tuning), averaged over the fine-tuning steps and normalised per pruning ratio. The selected rate, $3\cdot10^{-3}$, is used for all networks, pruning ratios and starting points.

\paragraph{Results.}
In JSD, the training objective, Magnitude + FT matches the one-shot \mthd edit after 15, 100 and 200 steps at 80\%, 50\% and 20\% remaining weights. At 20\% remaining weights, starting from the \mthd edit gives a lower JSD after 1500 steps on 93\% of the networks. Both curves are still decreasing at that point, and the gap narrows with more steps.

\subsection{Additional experiments}

\subsubsection{Ablation: Controlled experiments on ViT Zoo} On the CIFAR100 ViT zoo, the 3-lvl-H-GMN of Table~\ref{tab:gen_pred} has about $7.5\times$ more parameters than the
flat GMN (5.0M vs.\ 0.67M) and uses GRIT~\citep{ma2023graph} on its coarsest level, the Layer Graph. Its advantage
could therefore stem from its size or from the graph transformer rather than from the hierarchy. We control for both.

\textbf{Setup.} We fix the total number of trainable parameters to that of the flat GMN of
Table~\ref{tab:gen_pred}, a single GMN layer of width 52 on the Parameter Graph, which is the deepest flat GMN that
fits in our GPU memory. Each H-GMN applies one layer per level ($k_1=k_2=k_3=1$), on the Parameter, Head and Layer Graphs, with widths reduced to fit within the budget. The two variants differ only in $\mathfrak{F}_3$, the layer on the Layer Graph: a GMN layer, with width 35 at every level, or a GRIT layer with RRWP as in Table~\ref{tab:gen_pred}, with widths 34, 34 and 32. The pooling functions are those of the 3-lvl-H-GMN, and the readout and training recipe are shared with both models of Table~\ref{tab:gen_pred}. We train each variant with a single seed.

\textbf{Results.} As shown in Table~\ref{tab:gmn-grit-ablation}, both budget-matched H-GMNs outperform the flat GMN
and StatNN by a large margin. They also match the $\tau$ of the full-size 3-lvl-H-GMN of Table~\ref{tab:gen_pred},
with a lower $R^2$, using significantly fewer parameters, even with GMN on every level. Thus, the advantage of H-GMN
therefore does not come from its parameter count. Nor does it come from GRIT: the variant with GMN on every level
performs at least as well as the one with GRIT on the Layer Graph.

\begin{table}[h]
\centering
\caption{Generalisation prediction. 
We evaluate \mthd using both GMN and GRIT in the coarser levels, adhering to the parameter budget dictated by the best-performing GMN baseline. Rest of the baselines are reported same as in \cref{tab:gen_pred} 
We report Kendall's rank correlation ($\tau \uparrow$) and coefficient of determination ($R^2 \uparrow$).}
\label{tab:gmn-grit-ablation}
\begin{tabular}{@{}lccc@{}}
\toprule
Method & \multicolumn{3}{c}{CIFAR100 ViT} \\
\cmidrule(lr){2-4}
 & $\tau$ & $R^2$ & $\#\text{ Params}$ \\
\midrule
StatNN 
& $0.877$ & $0.811$ & $444,417$ \\
GMN 
& $0.804$ & $0.673$ & $665,177$\\
\midrule
\midrule
3-lvl-\mthdT (GMN-GMN-GMN) & $\mathbf{0.936}$ & $\mathbf{0.967}$ & $663,817$ \\
3-lvl-\mthdT (GMN-GMN-GRIT) & $0.924$ & $0.926$ & $663,479$ \\
\bottomrule
\end{tabular}
\end{table}

\subsubsection{Ablation: Deeper GMN on full Parameter Graph}\label{app:abl-deeper-mnist-vit}
H-GMN passes information between distant layers through its coarse graphs, whereas a flat GMN needs one message-passing layer per hop on the Parameter Graph. In this ablation, we test whether an increased receptive field of the GMN baseline (i.e., significantly increasing the GMN depth) could close the gap with our \mthd . On the CIFAR100 ViT zoo this cannot be tested, since each data-point model comprises of millions of edges. We therefore turn to the MNIST ViT zoo, whose Parameter Graphs have about a thousand times fewer edges (Table~\ref{tab:graph_stats}), so that flat GMNs with up to 16 layers, enough to span the graph's diameter, fit in memory.

\textbf{Setup.} We train flat GMNs with $k_1 \in \{2, 4, 6, 8, 16\}$ layers using the same configuration with the MNIST H-GMN (\cref{app:exp-details}): a hidden dimension of 128, LayerNorm, a learning rate of 0.0001, 200 epochs and an effective batch size of 64, using gradient accumulation where a full batch does not fit in GPU memory. Only the depth changes between runs. We compare them to the 3-lvl-H-GMN of \cref{tab:gen_pred}.

\begin{figure}[h!]
    \centering
    \includegraphics[width=0.6\textwidth]{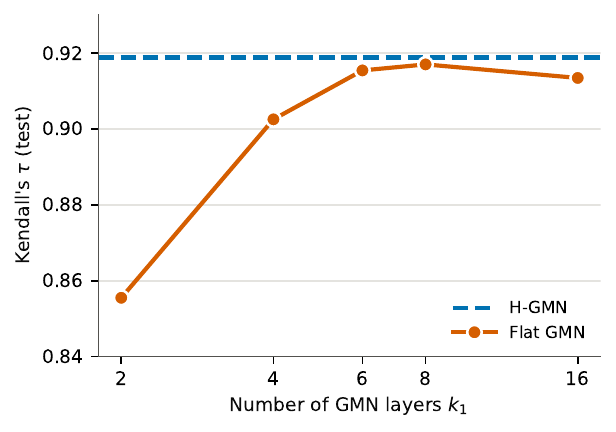}
    \caption{Flat GMN of increasing depth on the MNIST ViT zoo. Test Kendall's $\tau$ of flat GMNs with $k_1$ layers on the Parameter Graph, all trained with the configuration of the MNIST H-GMN, so that only the depth changes. Dashed: 3-lvl-H-GMN of Table~\ref{tab:gen_pred}. Eight layers suffice to match H-GMN; 16 layers, enough to span the diameter of the Parameter Graph, bring no further gain.}
    \label{fig:mnist-small-vit-gmn-depth}
\end{figure}

\textbf{Results.} As shown in \cref{fig:mnist-small-vit-gmn-depth}, depth closes the gap: going from 2 to 8 layers raises $\tau$ from $0.856$ to $0.916$, on par with H-GMN ($0.919$), and 16 layers bring no further gain. 
On this zoo, the advantage of H-GMN over a shallow flat GMN can be largely attributed to its larger receptive field, which a deep enough flat GMN can also reach, but at a much higher cost: the 8- and 16-layer flat GMNs took 8.1\,h and 15.9\,h to train on an RTX 4090, against 3.5\,h for H-GMN, while also requiring gradient accumulation to simulate the same batch size. 
On the CIFAR100 ViT zoo, where a single flat layer already occupies most of a 64\,GB GPU (\Cref{app:time-mem}), this route is closed, whereas H-GMN obtains its receptive field from the coarse graphs at little extra cost. 

\subsubsection{Varying ViT Zoo}

We further stress-test all metanetworks on a \textit{Beyond} split of 9 ViT architectures (360 networks), each larger than every training network: $60.3$–$85.9$M parameters versus at most $48.4$M in training, the largest being the unmodified ViT-B. Beyond differs from the training distribution in more than size. Its networks are also considerably more accurate (mean accuracy $0.83$ versus $0.55$ in training), and $64\%$ of them are more accurate than every network in the in-distribution test splits, so the metanetwork has to extrapolate in both input and label space. As shown in Table~\ref{tab:gen_pred_varying_vit_beyond}, all methods degrade. \mthd  is the only method with a positive $R^2$ on both Beyond splits, while StatNN retains a higher $\tau$.  The two methods fail differently: \mthd  systematically under-predicts the accuracy of the largest networks but keeps the spread of its predictions close to that of the labels, whereas StatNN ranks networks well but its predictions are over-dispersed, which yields a negative $R^2$. The ranking gap grows with the extrapolation distance. We treat Beyond as a stress test rather than a target setting; generalisation to architectures substantially larger than those seen in training is outside the scope of this work.

\begin{table}[h!]
\centering
\caption{Generalisation prediction on the Varying ViT zoo. We evaluate both on an in-distribution ($\leq 48 \text{M}$) test split and on a \textit{Beyond} test split ($60.3 \text{M} \leq \text{NN} \leq 85.9 \text{M}$). \textit{Beyond} $\leq 1.5\times$ keeps the Beyond architectures with at most $1.5\times$ the size of the largest training network.}
\label{tab:gen_pred_varying_vit_beyond}
\begin{tabular}{@{}lcccccc@{}}
\toprule
Method & \multicolumn{2}{c}{In-distribution Unseen} & \multicolumn{2}{c}{Beyond $\leq 1.5\times$} & \multicolumn{2}{c}{Beyond} \\
\cmidrule(lr){2-3} \cmidrule(lr){4-5} \cmidrule(lr){6-7}
 & $\tau$ & $R^2$ & $\tau$ & $R^2$ & $\tau$ & $R^2$ \\
\midrule
StatNN   & $0.795$ & $0.554$ & $0.481$ & $-0.481$ & $0.445$ & $-1.112$ \\
GMN      & $0.577$ & $0.491$ & $0.223$ & $-1.286$ & $0.079$ & $-2.444$ \\
\midrule
\midrule
\mthdT (Ours) & $0.849$ & $0.941$ & $0.324$ & $0.497$ & $0.207$ & $0.046$ \\
\bottomrule
\end{tabular}
\end{table}

\subsubsection{Generalizing on an unseen backbone}\label{app:lobo}
\paragraph{Generalisation to unseen pretraining backbones.}
Every model in the ViT zoo is fine-tuned from one of five pretrained backbones, and the random split places models from each backbone in both the training and test sets. A metanetwork can therefore score well by relating each test model to training models that share its initialization, which says little about whether it can handle weights from a pretraining run it has never seen. To test this, we perform a leave-one-backbone-out evaluation: we hold out all models fine-tuned from the sl backbone, one of the two backbones trained from scratch, and train on the remaining four. We evaluate on two splits: \textbf{test\_id}, the original test models from the four training backbones, and \textbf{held-out sl}, all models of the unseen backbone. Normalization statistics are computed on the reduced training set only, so no information about sl enters training, and all methods use the same hyperparameters as in the main experiments.

\paragraph{Results.}
Table~\ref{tab:eval-metrics-transposed} reports the results. On test\_id, H-GMN achieves the best overall and within-backbone metrics, and it is the only method that ranks the models of the cl backbone well, where StatNN and GMN fail. On the held-out sl backbone, performance drops sharply for all methods and $R^2$ is negative throughout, i.e.\ no method is more accurate than a constant predictor at the split's mean accuracy. H-GMN retains the strongest ranking ($\tau = 0.20$, versus $0.11$ for GMN and $\approx 0$ for StatNN), however generalisation across pretraining backbones remains an open challenge.

\begin{table*}[t]
\centering
\caption{Leave-one-backbone-out evaluation on the ViT-S/16 zoo: methods are trained on all backbones except sl. \textbf{test\_id}: held-out checkpoints from the
  training backbones; \textbf{held-out sl}: all checkpoints from the unseen sl backbone. We report Kendall's $\tau$ and $R^2$ (both $\uparrow$); wb denotes
  within-backbone metrics ($\tau_{\text{wb}}$: mean per-backbone $\tau$; $R^2_{\text{wb}}$: $R^2$ after removing each backbone's mean). Best results in bold.}
\label{tab:eval-metrics-transposed}
\resizebox{\textwidth}{!}{%
\begin{tabular}{@{}l cccc cc cc cc cc cc@{}}
\toprule
Method & \multicolumn{4}{c}{\textbf{test\_id --- overall}} & \multicolumn{2}{c}{\textbf{test\_id --- in1k}} & \multicolumn{2}{c}{\textbf{test\_id --- in21k}} & \multicolumn{2}{c}{\textbf{test\_id --- in21k\_ft\_in1k}} & \multicolumn{2}{c}{\textbf{test\_id --- cl}} & \multicolumn{2}{c}{\textbf{test --- held-out sl}} \\
\cmidrule(lr){2-5} \cmidrule(lr){6-7} \cmidrule(lr){8-9} \cmidrule(lr){10-11} \cmidrule(lr){12-13} \cmidrule(lr){14-15}
 & $\tau$ & $R^2$ & $\tau_{\text{wb}}$ & $R^2_{\text{wb}}$ & $\tau$ & $R^2$ & $\tau$ & $R^2$ & $\tau$ & $R^2$ & $\tau$ & $R^2$ & $\tau$ & $R^2$ \\
\midrule
StatNN 
 & $0.855$ & $0.790$ & $0.600$ & $0.804$ 
 & $\mathbf{0.882}$ & $0.964$ 
 & $0.785$ & $0.928$ 
 & $\mathbf{0.825}$ & $0.891$ 
 & $-0.092$ & $-0.235$ 
 & $0.003$ & $\mathbf{-0.048}$ \\
GMN 
 & $0.821$ & $0.783$ & $0.627$ & $0.767$ 
 & $0.826$ & $0.931$ 
 & $0.739$ & $0.978$ 
 & $0.785$ & $0.977$ 
 & $0.160$ & $-0.359$ 
 & $0.109$ & $-0.384$ \\
\midrule
\midrule
H-GMN 
 & $\mathbf{0.885}$ & $\mathbf{0.908}$ & $\mathbf{0.802}$ & $\mathbf{0.886}$ 
 & $0.835$ & $\mathbf{0.968}$ 
 & $\mathbf{0.829}$ & $\mathbf{0.985}$ 
 & $0.765$ & $\mathbf{0.989}$ 
 & $\mathbf{0.777}$ & $\mathbf{0.438}$ 
 & $\mathbf{0.203}$ & $-0.109$ \\
\bottomrule
\end{tabular}%
}
\end{table*}

\subsubsection{Ablation: H-GMN without fine-level processing}
\label{sec:no-fine-level}
In all other experiments, a single layer on the Parameter Graph ($k_1 = 1$) processes the input network before the first pooling. Here, we test how much of the performance of H-GMN can be attributed to the architecture-derived aggregation and to the subsequent processing of the coarse graphs, rather than to neuron-level processing or to more expressive coarse-level layers.

\paragraph{Setup.} We train a 3-lvl-H-GMN on generalisation prediction on the ResNet18 zoo, with no layer on the Parameter Graph ($k_1 = 0$): the vertex and edge embeddings produced by the graph initialisation are pooled directly into the Layer Graph, and then into the Block Graph, with no message passing on the Parameter Graph. All coarse-level modules are GMNs, so no graph transformer is used. The model has the same number of trainable parameters as the flat GMN of Table~\ref{tab:gen_pred}, and is trained on 10\%, 50\% and 100\% of the training set.

\paragraph{Results.} As shown in \Cref{tab:no-fine-level}, the variant without fine-level processing matches the full 3-lvl-H-GMN of Table~\ref{tab:gen_pred} at every training-set size ($\tau$ 0.851 / 0.882 / 0.898 vs.\ 0.850 / 0.881 / 0.893). It also clearly outperforms flat GMN in the low-data regime ($\tau$ 0.851 vs.\ 0.744 at 10\%) while matching it on the full training set, and it outperforms StatNN at every size. StatNN also summarises each layer with statistics, but without any message passing on top, so the gap between the two (e.g.\ $\tau$ 0.851 vs.\ 0.677 at 10\%) isolates the contribution of processing the architecture-derived coarse graphs. These results imply that on this zoo, aggregating along the architecture and reasoning over the resulting coarse graphs therefore accounts for most of the improvement of H-GMN, without neuron-level processing and without GRIT.

\begin{table}[h]
\centering
\caption{Generalisation prediction. We report Kendall's rank correlation ($\tau \uparrow$), coefficient of determination ($R^2 \uparrow$), and convergence epoch (Ep.). ``--'': architecturally inapplicable baseline.}
\label{tab:no-fine-level}
\resizebox{\textwidth}{!}{%
\begin{tabular}{@{}l*{3}{ccc}@{}}
\toprule
Method &
\multicolumn{3}{c}{ResNet18-0.1} &
\multicolumn{3}{c}{ResNet18-0.5} &
\multicolumn{3}{c}{ResNet18} \\
\cmidrule(lr){2-4} \cmidrule(lr){5-7} \cmidrule(lr){8-10}
 & $\tau$ & $R^2$ & Ep. & $\tau$ & $R^2$ & Ep. & $\tau$ & $R^2$ & Ep. \\
\midrule
StatNN 
& $0.677$ & $0.684$ & $92$ & $0.824$ & $0.884$ & $78$ & $0.837$ & $0.890$ & $82$ \\
GMN & $0.744$ & $0.840$ & $93$ & $0.886$ & $\mathbf{0.943}$ & $94$ & $0.900$ & $\mathbf{0.953}$ & $97$ \\
GMN (GRIT) 
& $0.764$ & $0.862$ & $96$ & $0.874$ & $0.924$ & $82$ & $0.889$ & $0.942$ & $91$ \\
\midrule
\midrule
2-lvl-\mthdT (Ours) & $0.846$ & $0.927$ & $80$ & $\mathbf{0.892}$ & $0.941$ & $74$ & $\mathbf{0.902}$ & $0.951$ & $82$ \\
2-lvl-\mthdT-\RC (Ours) & $0.837$ & $0.900$ & $81$ & $0.886$ & $0.938$ & $86$ & $0.900$ & $0.952$ & $85$ \\
3-lvl-\mthdT (Ours) & $0.850$ & $0.906$ & $\mathbf{61}$ & $0.881$ & $0.927$ & $\mathbf{63}$ & $0.893$ & $0.945$ & $\mathbf{50}$ \\
3-lvl-\mthdT-\RC (Ours) & $\mathbf{0.860}$ & $\mathbf{0.928}$ & $75$ & $0.882$ & $0.939$ & $65$ & $0.899$ & $0.942$ & $56$ \\
3-lvl-H-GMN ($k_1=0$) (Ours) & $0.851$ & $0.923$ & $83$ & $0.882$ & $0.931$ & $81$ & $0.898$ & $0.941$ & $84$ \\
\bottomrule
\end{tabular}%
}
\end{table}
\subsubsection{Multiple seeds on the backdoor model detection task}
All results of the backdoor model detection experiment (\Cref{tab:backdoor}) are averaged over three random seeds, except for the GMN (GRIT) baseline, for which only a single run was executed. To assess the robustness of these results, \Cref{tab:backdoor-seeds} reports again the mean of each metric over the three seeds, together with its standard deviation.

\begin{table}[h!]
\centering
\caption{Mean $\pm$ standard deviation over 3 seeds.}
\label{tab:backdoor-seeds}
\resizebox{\textwidth}{!}{%
\setlength{\tabcolsep}{4pt}
\begin{tabular}{@{}ll cccc@{}}
\toprule
Attack & Metric & StatNN & GMN & 2-lvl-\mthd (Ours) & 2-lvl-\mthd (GRIT) (Ours) \\
\midrule
\multirow{4}{*}{BadNets}
 & AUROC  & $0.986 \pm 0.002$ & $0.977 \pm 0.007$ & $0.896 \pm 0.122$ & $\mathbf{1.000 \pm 0.000}$ \\
 & Acc    & $0.971 \pm 0.001$ & $0.954 \pm 0.002$ & $0.913 \pm 0.053$ & $\mathbf{0.999 \pm 0.001}$ \\
 & F1     & $0.830 \pm 0.006$ & $0.767 \pm 0.028$ & $0.598 \pm 0.203$ & $\mathbf{0.996 \pm 0.006}$ \\
 & Recall & $0.775 \pm 0.009$ & $0.832 \pm 0.110$ & $0.664 \pm 0.189$ & $\mathbf{0.992 \pm 0.012}$ \\
\midrule
\multirow{4}{*}{Blended}
 & AUROC  & $0.999 \pm 0.001$ & $0.454 \pm 0.029$ & $\mathbf{1.000 \pm 0.000}$ & $0.999 \pm 0.001$ \\
 & Acc    & $0.991 \pm 0.005$ & $0.092 \pm 0.000$ & $0.995 \pm 0.005$ & $\mathbf{0.999 \pm 0.001}$ \\
 & F1     & $0.951 \pm 0.026$ & $0.170 \pm 0.000$ & $0.974 \pm 0.028$ & $\mathbf{0.981 \pm 0.015}$ \\
 & Recall & $0.978 \pm 0.017$ & $1.000 \pm 0.000$ & $0.970 \pm 0.029$ & $\mathbf{1.000 \pm 0.000}$ \\
\midrule
\multirow{4}{*}{WaNet}
 & AUROC  & $0.997 \pm 0.001$ & $0.553 \pm 0.053$ & $0.970 \pm 0.028$ & $\mathbf{1.000 \pm 0.000}$ \\
 & Acc    & $0.981 \pm 0.001$ & $0.503 \pm 0.368$ & $0.959 \pm 0.022$ & $\mathbf{0.999 \pm 0.001}$ \\
 & F1     & $0.896 \pm 0.011$ & $0.180 \pm 0.012$ & $0.775 \pm 0.108$ & $\mathbf{0.992 \pm 0.007}$ \\
 & Recall & $0.894 \pm 0.053$ & $0.566 \pm 0.388$ & $0.753 \pm 0.078$ & $\mathbf{0.995 \pm 0.009}$ \\
\midrule
\multirow{4}{*}{Multiclass}
 & AUROC  & $0.994 \pm 0.002$ & $0.959 \pm 0.001$ & $0.999 \pm 0.001$ & $\mathbf{0.999 \pm 0.002}$ \\
 & Acc    & $0.959 \pm 0.009$ & $0.891 \pm 0.011$ & $0.983 \pm 0.002$ & $\mathbf{0.984 \pm 0.001}$ \\
 & F1     & $0.923 \pm 0.019$ & $0.791 \pm 0.012$ & $0.950 \pm 0.009$ & $\mathbf{0.956 \pm 0.011}$ \\
 & Recall & $0.899 \pm 0.048$ & $0.783 \pm 0.006$ & $0.948 \pm 0.007$ & $\mathbf{0.955 \pm 0.005}$ \\
\bottomrule
\end{tabular}%
}
\end{table}

\subsection{Permutation equivariance of GMNs on MHSA parameter graphs}\label{sec:MHSA_perm}

We show that the graph construction of \cref{sec:graph-con} maps the permutation symmetries of an MHSA block to relabellings of the vertices and edges of its parameter graph, and that GMNs are equivariant to these relabellings, provided that positional encodings and learnable embeddings are shared across heads. 
As in \cref{sec:graph-con}, biases are omitted (see Remark~\ref{rem:mhsa_biases}).

\noindent\textbf{Parameter graph.} Consider an MHSA block $\fun$ with $\nheads$ heads and, for each head $h\in[\nheads]$, parameters $\bW^h_q,\bW^h_k\in\R^{d_k\times d_{\text{in}}}$, $\bW^h_v\in\R^{d_v\times d_{\text{in}}}$ and $\bW^h_o\in\R^{d_{\text{out}}\times d_v}$, all collected in $\bth$. Its parameter graph $\constr(\fun)$ has vertex set $\VP = \cV\cup\cV_o\cup\bigcup_{h}(\cV^h_k\cup\cV^h_v)$ (a disjoint union), edge set $\EP = \bigcup_{h}(\cE^h_k\cup\cE^h_v\cup\cE^h_{kv}\cup\cE^h_o)$, vertex features $\xv$, positional encodings $\pv,\pe$ and edge features
\begin{equation}\label{eq:mhsa_edges}
\xe(i,j) =
\begin{cases}
[\bW^h_q(j,i), \bW^h_k(j,i)], & (i,j)\in\cE^h_k = \cV\times\cV^h_k,\\
[\bW^h_v(j,i)], & (i,j)\in\cE^h_v = \cV\times\cV^h_v,\\
[1/\sqrt{d_k}], & (i,j)\in\cE^h_{kv} = \cV^h_k\times\cV^h_v,\\
[\bW^h_o(j,i)], & (i,j)\in\cE^h_o = \cV^h_v\times\cV_o,
\end{cases}
\qquad h\in[\nheads].
\end{equation}
Here, $(i,j)$ denotes an edge from $i$ to $j$ and, as in \cref{sec:graph-con}, each vertex is identified with its index within its vertex set (\eg $j\in\cV^h_k$ indexes a row of $\bW^h_q$). Only the features of $\cE^h_k$, $\cE^h_v$ and $\cE^h_o$ depend on $\bth$. All other attributes are \emph{$\bth$-independent}: the vertex features $\xv$, since all vertices are parameter-free and receive learnable embeddings (\cref{app:graph-constr}), the features of $\cE^h_{kv}$, and the positional encodings $\pv,\pe$, which include vertex and edge types. 

\noindent\textbf{Permutation Symmetries.} \cite{tran2025equivariant} studied the symmetries of Multi-Head Self-Attention modules and derived equivariant linear layers on their maximal symmetry group. Although self-attention parameters exhibit general linear symmetries, in this study, we focus on a subset thereof: permutation symmetries.

Every permutation $g = ((\pi^h_k,\pi^h_v)_{h=1}^{\nheads},\tau)\in\SthMHSA = (S_{d_k}\times S_{d_v})^{\nheads}\rtimes S_{\nheads}$, where $S_n$ the symmetric group over $n$ elements and $\rtimes$ the semi-direct product, transforms $\bth$ into $\bth' = g\cdot\bth$, such that the 
MHSA function remains unaltered $f_{C, \bth} = f_{C, g\cdot\bth}$.\footnote{Throughout this section, a prime marks the counterpart of a quantity, when transforming $\bth$ to $\bth'$; \eg $\xv'$ and $\xe'$ are the vertex and edge features of $\constr(f_{C,\bth'})$. } The element $g$ induces the map $\pi_g:\VP\to\VP$,
\begin{equation}\label{eq:mhsa_relabel}
\pi_g(i) =
\begin{cases}
i, & i\in\cV\cup\cV_o,\\
\text{the vertex of } \cV^{\tau(h)}_k \text{ with index } \pi^h_k(i), & i\in\cV^h_k,\\
\text{the vertex of } \cV^{\tau(h)}_v \text{ with index } \pi^h_v(i), & i\in\cV^h_v,
\end{cases}
\end{equation}
which fixes the input and output vertices, permutes the vertices within each head and relabels the heads by $\tau$; it is a bijection since $\tau$, $\pi^h_k$ and $\pi^h_v$ are bijections over their respective (non-overlapping) vertex sets.

To see why the function does not change, it is helpful to convert this into matrix form. The group acts on $\bth$ as follows: for all $h\in 1,\dots,\nheads$:
\begin{equation}\label{eq:mhsa_action}
\bW^{\prime h}_q = \bP^h_k\bW^{\tau(h)}_q,\quad
\bW^{\prime h}_k = \bP^h_k\bW^{\tau(h)}_k,\quad
\bW^{\prime h}_v = \bP^h_v\bW^{\tau(h)}_v,\quad
\bW^{\prime h}_o = \bW^{\tau(h)}_o(\bP^h_v)^\top,
\end{equation}
where $\bP^h_k,\bP^h_v$ are permutation matrices corresponding to,  $\pi^h_k,\pi^h_v$ (a permutation $\pi$ maps to a matrix $\bP$ with $\bP(a,\pi(a)) = 1$ and all other entries zero and thus $(\bP\bW)(a,i) = \bW(\pi(a),i)$ and $(\bW\bP^\top)(j,a) = \bW(j,\pi(a))$). Now, recall that  $(\bP^h_v)^\top\bP^h_v = (\bP^h_k)^\top\bP^h_k = \bI$. Thus head $h$ of $f_{C,\bth'}$ coincides with head $\tau(h)$ of $\fun$. Additionally, $f_{C,\bth'}$ is a summation over heads and since $\tau$ is a bijection, then $f_{C,\bth'} = \fun$. 

\begin{lemma}\label{lem:mhsa_graph}
Let $g\in\SthMHSA$ and $\bth' = g\cdot\bth$. Then:
(i) \emph{Edge set:} $(i,j)\mapsto(\pi_g(i),\pi_g(j))$ maps $\cE^h_k$, $\cE^h_v$, $\cE^h_{kv}$ and $\cE^h_o$ onto $\cE^{\tau(h)}_k$, $\cE^{\tau(h)}_v$, $\cE^{\tau(h)}_{kv}$ and $\cE^{\tau(h)}_o$, respectively, for all $h\in[\nheads]$; hence, $(i,j)\in\EP$ if and only if $(\pi_g(i),\pi_g(j))\in\EP$.
(ii) \emph{Edge features:} $\xe'(i,j) = \xe(\pi_g(i),\pi_g(j))$ for all $(i,j)\in\cE^h_k\cup\cE^h_v\cup\cE^h_o$ and $h\in[\nheads]$.
\end{lemma}

\begin{proof}
Write $\pi = \pi_g$. By \eqref{eq:mhsa_relabel}, $\pi$ is the identity on $\cV\cup\cV_o$ and restricts to bijections $\cV^h_k\to\cV^{\tau(h)}_k$ and $\cV^h_v\to\cV^{\tau(h)}_v$.
(i) Hence, $\pi$ maps $\cE^h_k = \cV\times\cV^h_k$ onto $\cV\times\cV^{\tau(h)}_k = \cE^{\tau(h)}_k$, and likewise for the other three edge sets. Since $\tau$ is a bijection, $\EP$ is mapped onto $\EP$; as $\pi$ is injective, $(\pi(i),\pi(j))\in\EP$ implies $(i,j)\in\EP$.
(ii) By \eqref{eq:mhsa_edges}--\eqref{eq:mhsa_relabel}, for $(i,j)$ in $\cE^h_k$, $\cE^h_v$ and $\cE^h_o$, respectively,
\begin{alignat*}{2}
\xe(\pi(i),\pi(j)) &= [\bW^{\tau(h)}_q(\pi^h_k(j),i),\bW^{\tau(h)}_k(\pi^h_k(j),i)] &\qquad& \text{for } (i,j)\in\cE^h_k,\\
&= [\bW^{\prime h}_q(j,i),\bW^{\prime h}_k(j,i)] = \xe'(i,j),\\
\xe(\pi(i),\pi(j)) &= [\bW^{\tau(h)}_v(\pi^h_v(j),i)] = [\bW^{\prime h}_v(j,i)] = \xe'(i,j), && \text{for } (i,j)\in\cE^h_v,\\
\xe(\pi(i),\pi(j)) &= [\bW^{\tau(h)}_o(j,\pi^h_v(i))] = [\bW^{\prime h}_o(j,i)] = \xe'(i,j), && \text{for } (i,j)\in\cE^h_o.
\end{alignat*}
\end{proof}

\begin{assumption}\label{ass:mhsa_attr}
For all $g\in\SthMHSA$, the $\bth$-independent attributes are invariant under $\pi_g$, \ie
\begin{alignat*}{2}
\xv(\pi_g(i)) &= \xv(i),\quad \pv(\pi_g(i)) = \pv(i), &\qquad& i\in\VP,\\
\pe(\pi_g(i),\pi_g(j)) &= \pe(i,j), && (i,j)\in\EP,\\
\xe(\pi_g(i),\pi_g(j)) &= \xe(i,j), && (i,j)\in\cE^h_{kv},\ h\in[\nheads].
\end{alignat*}
\end{assumption}
This holds whenever these attributes do not depend on the head index $h$ or on the position of a vertex within $\cV^h_k$ or $\cV^h_v$. This is the case for the type-based positional encodings and the layer-wise embeddings of \cref{app:graph-constr}, provided they are shared across heads, and it is automatic for the constant features of $\cE^h_{kv}$.

By Lemma~\ref{lem:mhsa_graph} and Assumption~\ref{ass:mhsa_attr}, $\constr(f_{C,\bth'})$ is thus $\constr(\fun)$ relabelled by $\pi_g$: $\pi_g$ preserves the common edge set $\EP$, and every attribute $a$ of $\constr(\fun)$ (\ie $\xv$, $\xe$, $\pv$ or $\pe$) and its counterpart $a'$ satisfy
\begin{equation}\label{eq:mhsa_input}
a'(i) = a(\pi_g(i))\ \ \forall i\in\VP, \qquad a'(i,j) = a(\pi_g(i),\pi_g(j))\ \ \forall (i,j)\in\EP.
\end{equation}
Indeed, Lemma~\ref{lem:mhsa_graph}(ii) covers $\xe$ on $\cE^h_k$, $\cE^h_v$ and $\cE^h_o$; all other attributes are $\bth$-independent, so that $a' = a$ and \eqref{eq:mhsa_input} is Assumption~\ref{ass:mhsa_attr}.

\noindent\textbf{GMN.} We consider GMNs with forward and backward messages, edge updates and positional encodings \citep{lim2024graph,kofinas2024graph,kalogeropoulos2024scale}. With $\cN_F(i) = \{j : (j,i)\in\EP\}$, $\cN_B(i) = \{j : (i,j)\in\EP\}$ and $t = 1,\dots,T$, for $i\in\VP$ and $(i,j)\in\EP$,
\begin{equation}\label{eq:mhsa_gmn}
\begin{aligned}
\hv^0(i) &= \INIT_V\big(\xv(i),\pv(i)\big), \qquad \he^0(i,j) = \INIT_E\big(\xe(i,j),\pe(i,j)\big),\\
\mvfw^t(i) &= \textstyle\bigoplus_{j\in\cN_F(i)}\MSG^t_F\big(\hv^{t-1}(i),\hv^{t-1}(j),\he^{t-1}(j,i),\pv(i),\pv(j),\pe(j,i)\big),\\
\mvbw^t(i) &= \textstyle\bigoplus_{j\in\cN_B(i)}\MSG^t_B\big(\hv^{t-1}(i),\hv^{t-1}(j),\he^{t-1}(i,j),\pv(i),\pv(j),\pe(i,j)\big),\\
\hv^t(i) &= \UPD^t_V\big(\hv^{t-1}(i),\mvfw^t(i),\mvbw^t(i),\pv(i)\big),\\
\he^t(i,j) &= \UPD^t_E\big(\hv^{t-1}(i),\hv^{t-1}(j),\he^{t-1}(i,j),\pv(i),\pv(j),\pe(i,j)\big),
\end{aligned}
\end{equation}
where $\bigoplus$ is a permutation-invariant aggregator, \ie a function of the multiset of its arguments (\eg sum, mean, max or PNA), and $\INIT$, $\MSG$ and $\UPD$ are arbitrary functions (\eg MLPs) shared across vertices and edges. Single-direction GMNs are obtained by dropping $\mvbw^t$, and edge types enter through $\pe$. For invariant tasks, a permutation-invariant readout $\READ\big(\{\!\{\hv^T(i) : i\in\VP\}\!\}\big)$ follows.

\begin{proposition}\label{prop:mhsa_equiv}
Under Assumption \ref{ass:mhsa_attr}, let $g\in\SthMHSA$, $\bth' = g\cdot\bth$, and let $(\hv^t,\he^t)$ and $(\hv^{\prime t},\he^{\prime t})$ be the representations computed by the same GMN at layer $t$ on $\constr(\fun)$ and $\constr(f_{C,\bth'})$, respectively. Then, for all $t\in\{0,\dots,T\}$,
\begin{equation}\label{eq:mhsa_equiv}
\hv^{\prime t}(i) = \hv^t(\pi_g(i))\ \ \forall i\in\VP, \qquad \he^{\prime t}(i,j) = \he^t(\pi_g(i),\pi_g(j))\ \ \forall (i,j)\in\EP.
\end{equation}
By \eqref{eq:mhsa_input}, transforming the network by $g$ relabels its parameter graph by $\pi_g$, and \eqref{eq:mhsa_equiv} states that the representations computed by the GMN are relabelled in the same way. Hence, the GMN is $\SthMHSA$-equivariant, with $g$ acting on parameters by $\bth\mapsto g\cdot\bth$ and on vertex and edge representations by relabelling with $\pi_g$; followed by a permutation-invariant readout, it is $\SthMHSA$-invariant.
\end{proposition}
\begin{proof}
Fix $g$ and write $\pi = \pi_g$. By Lemma~\ref{lem:mhsa_graph}(i), $j\in\cN_F(i)\iff(j,i)\in\EP\iff(\pi(j),\pi(i))\in\EP\iff\pi(j)\in\cN_F(\pi(i))$, so, as $\pi$ is a bijection of $\VP$, it restricts to a bijection $\cN_F(i)\to\cN_F(\pi(i))$ and, likewise, $\cN_B(i)\to\cN_B(\pi(i))$, for every $i\in\VP$. By \eqref{eq:mhsa_input}, the positional encodings, which are the same for $\bth$ and $\bth'$, satisfy $\pv(i) = \pv(\pi(i))$ and $\pe(i,j) = \pe(\pi(i),\pi(j))$, \ie the same identities as \eqref{eq:mhsa_equiv}. As they enter \eqref{eq:mhsa_gmn} in the same way as $\hv^{t-1}$ and $\he^{t-1}$, we omit them from the arguments below. We proceed by induction on $t$.

\emph{Base case ($t=0$).} By \eqref{eq:mhsa_input}, $\xv'(i) = \xv(\pi(i))$ and $\xe'(i,j) = \xe(\pi(i),\pi(j))$. Hence,
\begin{align*}
\hv^{\prime 0}(i) &= \INIT_V\big(\xv'(i)\big) = \INIT_V\big(\xv(\pi(i))\big) = \hv^0(\pi(i)),\\
\he^{\prime 0}(i,j) &= \INIT_E\big(\xe'(i,j)\big) = \INIT_E\big(\xe(\pi(i),\pi(j))\big) = \he^0(\pi(i),\pi(j)).
\end{align*}

\emph{Induction step.} Suppose \eqref{eq:mhsa_equiv} holds for $t-1$. For the forward messages,
\begin{align*}
\mvfw^{\prime t}(i)
&= \textstyle\bigoplus_{j\in\cN_F(i)}\MSG^t_F\big(\hv^{\prime t-1}(i),\hv^{\prime t-1}(j),\he^{\prime t-1}(j,i)\big)\\
&= \textstyle\bigoplus_{j\in\cN_F(i)}\MSG^t_F\big(\hv^{t-1}(\pi(i)),\hv^{t-1}(\pi(j)),\he^{t-1}(\pi(j),\pi(i))\big)\\
&= \textstyle\bigoplus_{j\in\cN_F(\pi(i))}\MSG^t_F\big(\hv^{t-1}(\pi(i)),\hv^{t-1}(j),\he^{t-1}(j,\pi(i))\big) = \mvfw^t(\pi(i)),
\end{align*}
where the second equality is the induction hypothesis and the third holds because $\pi$ is a bijection from $\cN_F(i)$ to $\cN_F(\pi(i))$, so both aggregations are over the same multiset. The same argument with $\cN_B$ gives $\mvbw^{\prime t}(i) = \mvbw^t(\pi(i))$. Therefore,
\begin{align*}
\hv^{\prime t}(i) &= \UPD^t_V\big(\hv^{\prime t-1}(i),\mvfw^{\prime t}(i),\mvbw^{\prime t}(i)\big)\\
&= \UPD^t_V\big(\hv^{t-1}(\pi(i)),\mvfw^t(\pi(i)),\mvbw^t(\pi(i))\big) = \hv^t(\pi(i)),\\
\he^{\prime t}(i,j) &= \UPD^t_E\big(\hv^{\prime t-1}(i),\hv^{\prime t-1}(j),\he^{\prime t-1}(i,j)\big)\\
&= \UPD^t_E\big(\hv^{t-1}(\pi(i)),\hv^{t-1}(\pi(j)),\he^{t-1}(\pi(i),\pi(j))\big) = \he^t(\pi(i),\pi(j)),
\end{align*}
where $(\pi(i),\pi(j))\in\EP$ by Lemma~\ref{lem:mhsa_graph}(i). This completes the induction. Finally, since $\pi$ is a bijection of $\VP$, the multisets $\{\!\{\hv^{\prime T}(i) : i\in\VP\}\!\}$ and $\{\!\{\hv^T(i) : i\in\VP\}\!\}$ coincide, and so do the outputs of any permutation-invariant readout.
\end{proof}

\begin{remark}\label{rem:mhsa_biases}
With biases $\bb^h_q,\bb^h_k\in\R^{d_k}$, $\bb^h_v\in\R^{d_v}$ and $\bb_o\in\R^{d_{\text{out}}}$, we set $\xv(j) = [\bb^h_q(j),\bb^h_k(j)]$ for $j\in\cV^h_k$, $\xv(j) = [\bb^h_v(j)]$ for $j\in\cV^h_v$ and $\xv(j) = [\bb_o(j)]$ for $j\in\cV_o$, and $g$ transforms the biases as $\bb^{\prime h}_q = \bP^h_k\bb^{\tau(h)}_q$, $\bb^{\prime h}_k = \bP^h_k\bb^{\tau(h)}_k$, $\bb^{\prime h}_v = \bP^h_v\bb^{\tau(h)}_v$ and $\bb'_o = \bb_o$, which again preserves $\fun$. As in Lemma~\ref{lem:mhsa_graph}(ii), this gives $\xv'(i) = \xv(\pi_g(i))$ on $\cV_o$ and on all $\cV^h_k$, $\cV^h_v$, so the condition on $\xv$ in Assumption~\ref{ass:mhsa_attr} is only needed on $\cV$, where it holds trivially. Hence, \eqref{eq:mhsa_input} still holds, and so does Proposition~\ref{prop:mhsa_equiv}, with the same proof. Moreover, since $\cV^h_k$ and $\cV^h_v$ are only incident to the edges in \eqref{eq:mhsa_edges}, the proof of Proposition~\ref{prop:mhsa_equiv} applies when the block is part of a larger parameter graph, with $\pi_g$ extended by the identity to all other vertices.
\end{remark}

\end{document}